\documentclass[10pt,twocolumn,letterpaper]{article}

\usepackage[pagenumbers]{cvpr} % To force page numbers, e.g. for an arXiv version

\usepackage{times}
\usepackage{epsfig}
\usepackage{graphicx}
\usepackage{amsmath}
\usepackage{amssymb}
\usepackage{booktabs}
\usepackage{tabularx} 
\usepackage[dvipsnames]{xcolor}
\usepackage{xcolor}
\usepackage{color}
\usepackage{colortbl}
\usepackage{pifont}
\usepackage{multirow}
\usepackage{makecell}

\usepackage{lipsum}

\newcommand\blfootnote[1]{%
  \begingroup
  \renewcommand\thefootnote{}\footnote{#1}%
  \addtocounter{footnote}{-1}%
  \endgroup
}

\definecolor{turquoise}{cmyk}{0.65,0,0.1,0.3}
\definecolor{purple}{rgb}{0.65,0,0.65}
\definecolor{dark_green}{rgb}{0, 0.5, 0}
\definecolor{orange}{rgb}{0.8, 0.6, 0.2}
\definecolor{red}{rgb}{0.8, 0.2, 0.2}
\definecolor{darkred}{rgb}{0.6, 0.1, 0.05}
\definecolor{blueish}{rgb}{0.0, 0.3, .6}
\definecolor{blue}{rgb}{0, 0.3, 1}
\definecolor{light_gray}{rgb}{0.7, 0.7, .7}
\definecolor{pink}{rgb}{1, 0, 1}
\definecolor{greyblue}{rgb}{0.25, 0.25, 1}
\definecolor{light_gray}{gray}{0.95}
\definecolor{light-green}{rgb}{0.82, 0.94, 0.75}

\newcommand{\hz}{\vphantom{\parbox[c]{0.25cm}{\rule{0.25cm}{0.28cm}}}}
\newcommand{\frozen}[1]{{\setlength\fboxsep{1.5pt}\colorbox{blue!20}{\hz{$\displaystyle #1$}}}}
\newcommand{\learn}[1]{{\setlength\fboxsep{1.5pt}\colorbox{red!20}{\hz{$\displaystyle #1$}}}}

\renewcommand{\paragraph}[1]{\vspace{1em}\noindent\textbf{#1}}

\newcommand{\cls}{\texttt{[CLS]}}
\newcommand{\bp}{\mathbf{p}}
\newcommand{\bpsp}{\mathbf{p}^{\text{SP}}}
\newcommand{\bptp}{\mathbf{p}^{\text{TP}}}
\newcommand{\bh}{\mathbf{h}}
\newcommand{\be}{\mathbf{e}}
\newcommand{\bz}{\mathbf{z}}
\newcommand{\bx}{\mathbf{x}}
\newcommand{\by}{\mathbf{y}}
\newcommand{\bv}{\mathbf{v}}
\newcommand{\spcls}{\bx_t^{\text{SP}}\{\cls\}}
\newcommand{\tpcls}{\bx_g^{\text{TP}}\{\cls\}}

\newcommand{\mlp}{\texttt{MLP}}
\newcommand{\lnorm}{\texttt{LN}}
\newcommand{\mha}{\texttt{MHA}}
\newcommand{\reshape}{\texttt{Reshape}}
\newcommand{\conv}{\texttt{Conv2d}}
\newcommand{\dwconv}{\texttt{D}\text{-}\texttt{Conv3d}}

\newcommand{\paragrapht}[1]{\noindent\textbf{#1}}  % tidy \paragraph

\newcommand{\xmark}{\ding{55}}%

\newcommand{\method}{\textsc{DualPath}\xspace}

\usepackage[dvipsnames]{xcolor}
\usepackage{xcolor,colortbl}
\usepackage{pythonhighlight}
\definecolor{F7E0D5}{RGB}{247,224,213}
\colorlet{Light}{White!0!F7E0D5}

\usepackage[font=small]{caption}
\usepackage{mathtools}
\usepackage{graphicx}
\usepackage{soul}
\usepackage{enumitem}

\usepackage{xspace}
\usepackage{multirow}
\usepackage{array}
\usepackage{layouts}
\usepackage{algorithm}
\usepackage{bbding}
\usepackage{listings}
\usepackage{pifont}
\usepackage{wasysym}
\usepackage{float}
\usepackage{soul}
\usepackage{footnote}
\makesavenoteenv{tabular}
\makesavenoteenv{table}
\makesavenoteenv{figure}
\usepackage{pythonhighlight}

\newcommand{\Eq}[1]{Eq.~\ref{equ:#1}}

\usepackage[pagebackref,breaklinks,colorlinks]{hyperref}
 \makeatletter
 \def\hlinewd#1{%
      \noalign{\ifnum0=`}\fi\hrule \@height #1 \futurelet
      \reserved@a\@xhline}

\usepackage[pagebackref,breaklinks,colorlinks]{hyperref}

\usepackage[capitalize]{cleveref}
\crefname{section}{Sec.}{Secs.}
\Crefname{section}{Section}{Sections}
\Crefname{table}{Table}{Tables}
\crefname{table}{Tab.}{Tabs.}

\def\cvprPaperID{10456} % *** Enter the CVPR Paper ID here
\def\confName{CVPR}
\def\confYear{2023}
\newcommand{\printfnsymbol}[1]{%
  \textsuperscript{\@fnsymbol{#1}}%
}

\begin{document}

\title{Dual-path Adaptation from Image to Video Transformers}

\author{Jungin Park$^{1*}$
% \thanks{Both authors contributed equally to this work.} 
\quad\quad\quad Jiyoung Lee$^{2*}$ 
\quad\quad\quad Kwanghoon Sohn$^{1,3\dagger}$
% \thanks{Corresponding author.} 
\vspace{5pt}\\
$^1$Yonsei University \quad\quad $^2$NAVER AI Lab \quad\quad $^3$Korea Institute of Science and Technology (KIST)\vspace{3pt}\\
{\tt\small $\lbrace$newrun, khsohn$\rbrace$@yonsei.ac.kr} \quad\quad\quad
\tt\small lee.j@navercorp.com}

\maketitle

%%%%%%%%% ABSTRACT
\begin{abstract}
    \blfootnote{\hskip -0.2in $*$ Equal contributions. \quad $\dagger$ Corresponding author.}
    \blfootnote{\hskip -0.2in Official code: \url{https://github.com/park-jungin/DualPath}}
   In this paper, we efficiently transfer the surpassing representation power of the vision foundation models, such as ViT and Swin, for video understanding with only a few trainable parameters.
   Previous adaptation methods have simultaneously considered spatial and temporal modeling with a unified learnable module but still suffered from fully leveraging the representative capabilities of image transformers.
   We argue that the popular dual-path (two-stream) architecture in video models can mitigate this problem.
   We propose a novel \method adaptation separated into spatial and temporal adaptation paths, where a lightweight bottleneck adapter is employed in each transformer block.
   Especially for temporal dynamic modeling, we incorporate consecutive frames into a grid-like frameset to precisely imitate vision transformers' capability that extrapolates relationships between tokens.
   In addition, we extensively investigate the multiple baselines from a unified perspective in video understanding and compare them with \method.
   Experimental results on four action recognition benchmarks prove that pretrained image transformers with \method can be effectively generalized beyond the data domain.
   \vspace{-5pt}
\end{abstract}
\vspace{-5pt}
\section{Introduction}
\label{sec:intro}
\vspace{-5pt}
    Recognizing \textit{when}, \textit{where}, and \textit{what} happened is a fundamental capability in the human cognition system to understand our natural world.
    The research for video understanding inspires such capability for machine intelligence to comprehend scenes over time flow.
    Over the last decade, the development of deep neural networks~\cite{c3d, i3d, s3d, r(2+1)d} has contributed towards advances in video understanding.

    Vision Transformer (ViT)~\cite{vit} has recently emerged, making an upheaval in the research field of computer vision.
    ViT and its variants~\cite{swin, cvt, tokens-to-tokens, cswin} have demonstrated remarkable generalizability and transferability of their representations with scaled-up foundation models~\cite{clip, align, florence, beit-pretraining, pretraining1, pretraining2} and large-scale web-collected image data (\eg JFT-3B~\cite{pretraining2}, LAION-5B~\cite{laion-5b}).
    To capitalize on well-trained visual foundation models, finetuning entire parameters of the pretrained models with task-specific objectives has been the most popular transfer technique.
    However, it requires high-quality training data and plenty of computational resources to update the whole parameters for each downstream task, making overwhelming efforts for training.
    While partial finetuning~\cite{moco}, which trains additional multilayer perceptron (MLP) layers to the top of the model, has also been widely used for affordable training costs, unsatisfactory performance has been pointed out as a problem.

    \begin{figure}[t]
	\centering
	\includegraphics[width=0.99 \linewidth]{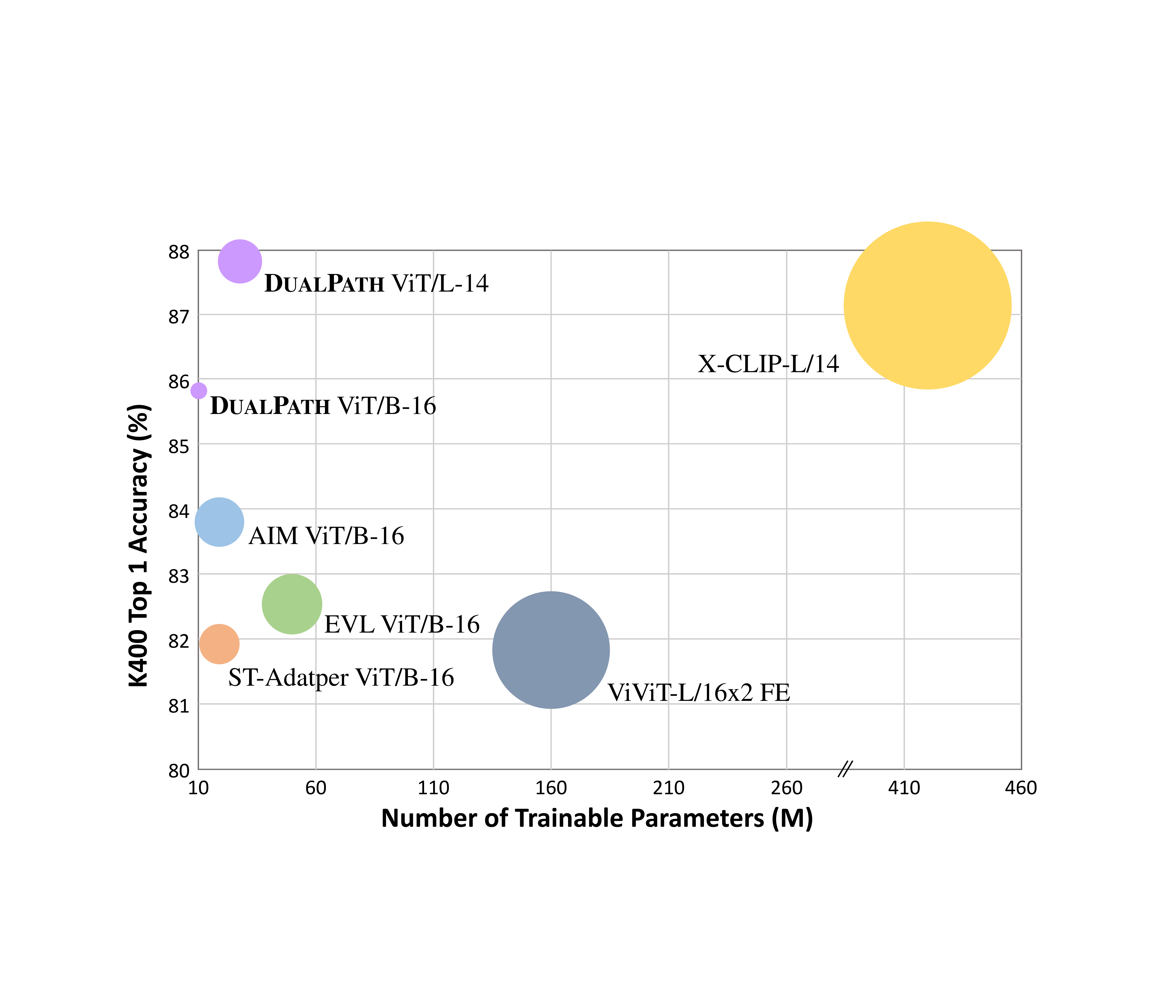}\\ \vspace{-7pt}
	\caption{Performance comparison on the Kinetics-400~\cite{k400} dataset. We depict the action recognition performance (vertical axis, \%) with respect to the number of trainable parameters (horizontal axis). The size of circles indicates GFLOPs for inference.}\vspace{-5pt}
	\label{fig:1}
    \end{figure}
    
    Most recently, parameter-efficient transfer learning (PETL) methods~\cite{adapter-nlp1, adapter-nlp2, prefix, prompt-nlp, diff, lora} have been proposed as an alternative to finetuning in the natural language processing area to adapt the large-scale language model, such as GPT series~\cite{gpt-1, gpt-2, gpt-3} and T5~\cite{t5}, for each task.
    They have successfully attained comparable or even surpassing performance to full-tuning parameters by learning a small number of extra trainable parameters only while keeping the original parameters of the pretrained model frozen.
    Thanks to their effectiveness and simplicity, they have been extended to vision models by applying prompt-based methods~\cite{vpt, visual-prompt} and adapter-based methods~\cite{adaptformer, protuning, vl-adapter}.
    They have efficiently adapted pretrained models to downstream tasks with significantly reduced tuning parameters, but most of these works mainly focus on transferring image models to image tasks~\cite{vpt, visual-prompt, adaptformer, protuning} and vision-language models to vision-language tasks~\cite{vl-adapter}.
    Inspired by the advances of the prior arts, we raise two conceivable questions: \textbf{(1)} Is it possible to transfer the parameters of the image foundation model to another video domain? \textbf{(2)} Is it also possible the transferred model performs comparably to the carefully designed video models that take the spatiotemporal nature of the video into account?

    While image models have demonstrated strong spatial context modeling capabilities~\cite{clip, align, pretraining1, pretraining2}, video transformer models~\cite{vivit,video-swin,tokenlearner,multiview} require a more complex architecture (\eg 539 vs 48912 GFLOPs~\cite{tokenlearner}) with a large number of parameters (\eg 84M vs 876M parameters~\cite{multiview}) than ViT for temporal context reasoning.
    Therefore, the challenge in transferring image models for video understanding is to encode the temporal context of videos while leveraging the discriminative spatial context of the pretrained image models.
    A naive solution is to finetune image models on a video dataset by directly applying previous prompt-/adapter-based approaches~\cite{vpt, visual-prompt, adaptformer, protuning}.
    However, these approaches inevitably ignore the temporal context in videos because they bridge only the spatial contexts between image and video data.

    In this paper, we propose a novel adapter-based dual-path parameter efficient tuning method for video understanding, namely \textbf{\method}, which consists of two distinct paths (\textit{spatial} path and \textit{temporal} path).
    For both paths, we freeze the pretrained image model and train only additional bottleneck adapters for tuning.
    The \textbf{spatial path} is designed to encode the spatial contexts that can be inferred from the appearance of individual frames with the minimum tuning of the pretrained image model.
    To reduce the computation burden, we sparsely use the frames with a low frame rate in the spatial path. 
    The \textbf{temporal path} corresponds to the temporal context that should be encoded by grasping the dynamic relationship over several frames sampled with a high frame rate.
    Especially for two reasons, we construct a grid-like frameset that consists of consecutive low-resolution frames as an input of the temporal path: (i) preventing computational efficiency loss caused by calculating multiple frames simultaneously; (ii) precisely imitating the ViT's ability for extrapolating global dependencies between input tokens.
    To compare our \method with existing methods broadly, we implement several baselines with a unified perspective on recent domain-specific PETL approaches~\cite{adaptformer, vpt, protuning}.
    Extensive experiments on several action recognition benchmarks~\cite{k400, hmdb51, ssv2, diving} demonstrate the effectiveness and high efficiency of our \method, achieving comparable and even better performance than the baselines and prior video models~\cite{multiscale, multiscale-v2, uniformer, space-time, vivit, video-swin, multiview, tokenlearner, actionclip, omnivore}.
    We achieve these results with extremely low computational costs for both training and inference, as demonstrated in \figref{fig:1}.

\section{Related Work}
\vspace{-5pt}
\paragrapht{Pretraining vision models.}
    To address the burdens of collecting large-scale labeled datasets for supervised learning~\cite{imagenet, pretraining1, pretraining2}, self-supervised learning methods~\cite{moco, simclr, byol, mae, empirical, simmim} have been introduced to learn general-purpose visual representations from unlabeled data.
    Similarly, self-supervised learning methods for videos have also been proposed with large-scale unlabeled video/video-language data~\cite{video-moco, provico, mae-st, videomae,videoclip, actionclip, videobert}.
    However, collecting even unlabeled video-language pairs is still quite costly compared to image-language pairs.
    In addition, pretraining video models require more computational power than images.
    We thus take advantage of the powerful pretrained image-based models for efficient video understanding.

\paragrapht{Video action recognition.}
    Action recognition is one of the most fundamental research topics for video understanding.
    Early works have been built upon convolution neural networks (CNNs)~\cite{i3d, s3d, slowfast, tsm, r(2+1)d} to effectively infer the spatiotemporal context for action recognition.
    Since Vision Transformer (ViT)~\cite{vit} has become a new paradigm in computer vision, transformers for video understanding have been actively studied by extending pretrained image models.
    The pretrained image transformers have been used to initialize the part of the video transformers~\cite{vivit, space-time, vidtr, multiview} or inflated to the video transformers~\cite{video-swin}.
    While transformers have demonstrated superior performance on video action recognition, they require full finetuning on video datasets, making the training inefficient.

    \paragrapht{Parameter-efficient transfer learning (PETL).}
    To address the memory and parameter inefficiency of full-/partial-finetuning, PETL has first introduced in natural language processing (NLP)~\cite{adapter-nlp1, adapter-nlp2, diff, lora, prompt-nlp, prefix, bitfit}.
    The main objective of PETL is to attain comparable or surpassing performance on downstream tasks by finetuning with only a small number of trainable parameters.
    Although PETL approaches~\cite{vpt, adaptformer, visual-prompt, bypass, protuning, vl-adapter, adaptformer} have recently been studied in computer vision, they are `blind' to other modalities such that image models are used for image tasks, and so are the other modalities.
    In contrast, we share the same objective as recent works for image-to-video transfer learning~\cite{frozen-clip, x-clip, st-adapter, videoprompt}, demonstrating the pretrained image models can be good video learners.
    However, they have several limitations in terms of parameter and computational efficiency.
    For example, \cite{frozen-clip} learned an extra decoder that contains 3D convolution layers and cross-frame attention, and \cite{st-adapter} inserted additional depth-wise 3D convolution layers between the down-/up-projection layers of the adapter to perform temporal reasoning, inducing computational inefficiency.
    The most recent works~\cite{videoprompt, x-clip} require an additional text encoder branch as a classifier.
    Moreover, they have computational efficiency proportional to the temporal resolution.
    Our \method accomplishes more efficient spatiotemporal modeling while achieving higher performance.
    
% \vspace{-5pt}
\section{Preliminaries and Baselines}\label{sec:baseline}
\vspace{-3pt}
    \subsection{Vision transformers for video}\vspace{-5pt}
    We briefly describe how to apply vision transformers for video understanding below.
    Following \cite{transformer}, given a set of $T$ frames in a video, we split each frame into $N$ patches of size $(P \times P)$ and tokenize them using a linear projection, such that
    \begin{equation}\label{equ:tokenization}
        {\mathbf{X}}_t = [\bx_t\{\cls\}, {\bx}_{t}^{1}, {\bx}_{t}^{2}, \cdots, {\bx}_{t}^{N}] + \bp,
    \end{equation}
    where ${\mathbf{X}}_t$ is a set of tokens for the $t$-th frame, and $\bx_t\{\cls\}$ and $\bp$ denote a learnable class token and a learned positional embedding respectively.
    We feed $(N+1)$ tokens of each frame to a sequence of $L$ transformer blocks, and the output of the $l$-th block $\bh_{l, t}$ can be derived by the following equations:
    \begin{equation}
    \begin{aligned}
        \bz_{l, t} = \bh_{l-1, t} &+ \mha(\lnorm(\bh_{l-1, t})),  \\
        \bh_{l, t} = \bz_{l, t} &+ \mlp(\lnorm(\bz_{l, t})),
    \end{aligned}\label{equ:transformer}
    \end{equation}
    where $\lnorm, \mha$, and $\mlp$ denote a layer normalization~\cite{layernorm}, multi-head attention~\cite{transformer}, and a multilayer perceptron operation, respectively.
    We apply layer normalization to the learned $T$ class tokens from the final transformer block and treat them as a set of frame representations.

    To take minimal temporal modeling into account the following baselines~\cite{vpt, adaptformer, protuning}, we employ a temporal transformer block followed by a full-connected (FC) layer as a classifier for video action recognition, similar to \cite{videoprompt}.
    We add learnable temporal positional embeddings $\bp_\text{temp}$ to the frame representations (\ie, $\bx_t\{\texttt{[cls]}\} \leftarrow \bx_t\{\texttt{[cls]}\} + \bp_\text{temp}$) and feed them into the transformer classifier.
    For the ST-adapter~\cite{st-adapter}, we use a single FC layer as a classifier.

     \begin{figure}[t]
        \centering
            \begin{subfigure}[t]{0.45\linewidth}
            \centering
            {\includegraphics[width=0.95\linewidth]{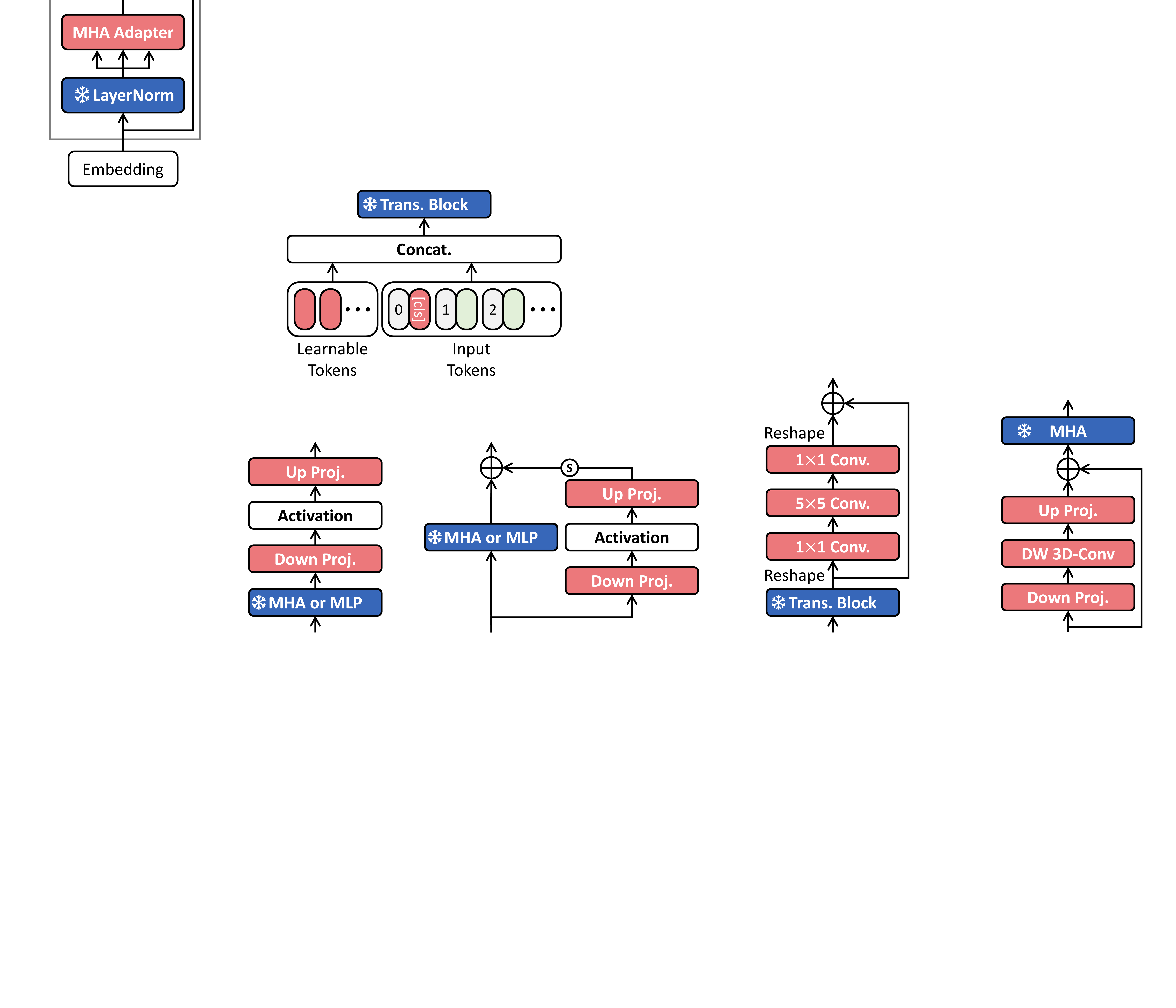}}
            \caption{VPT~\cite{vpt}}\label{fig:2a}
           \end{subfigure}
           \begin{subfigure}[t]{0.45\linewidth}
            \centering
            {\includegraphics[width=0.95\linewidth]{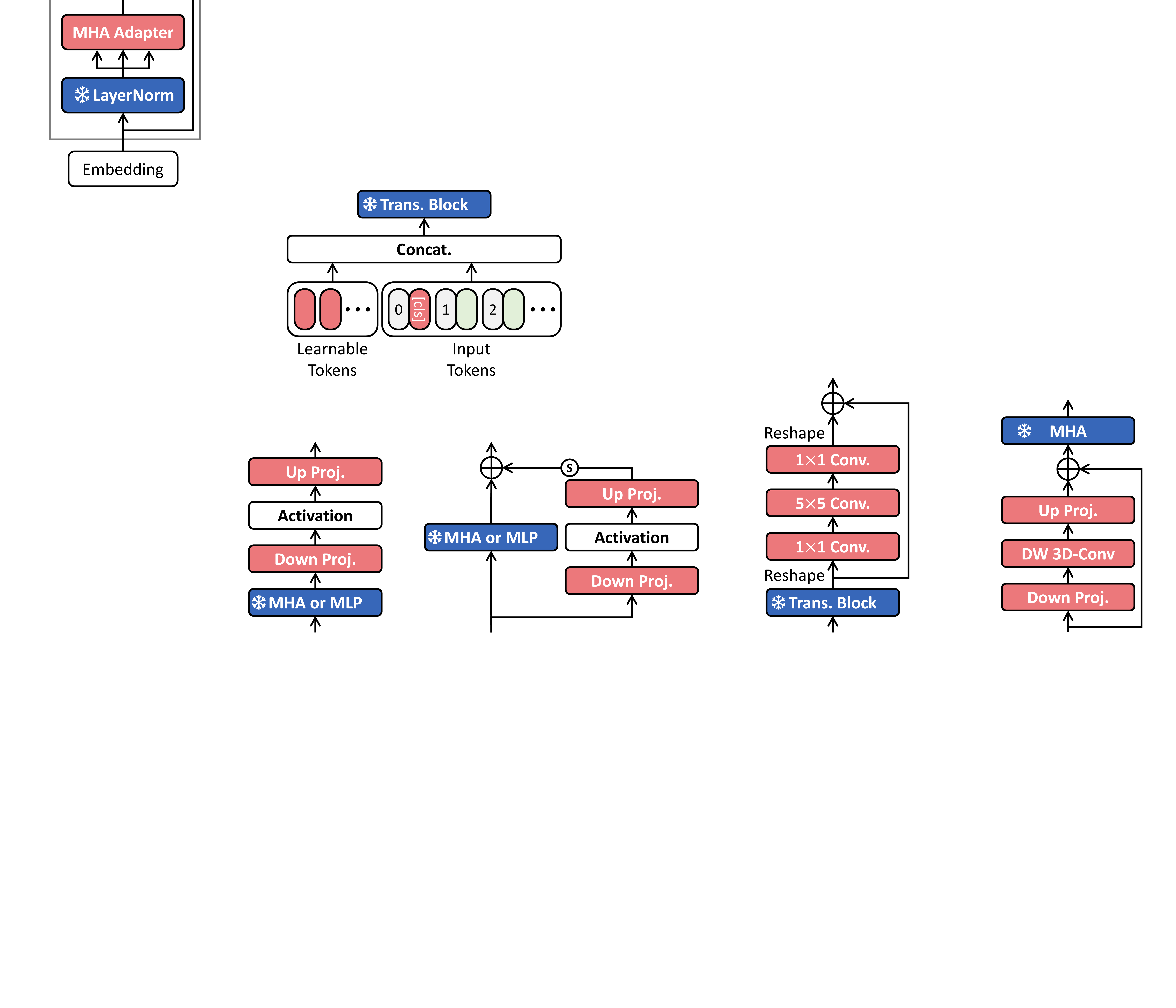}}
            \caption{{Parallel Adapter}~\cite{adaptformer}}\label{fig:2b}
           \end{subfigure}  \\
           \begin{subfigure}[t]{0.32\linewidth}
            \centering
            {\includegraphics[width=0.7\linewidth]{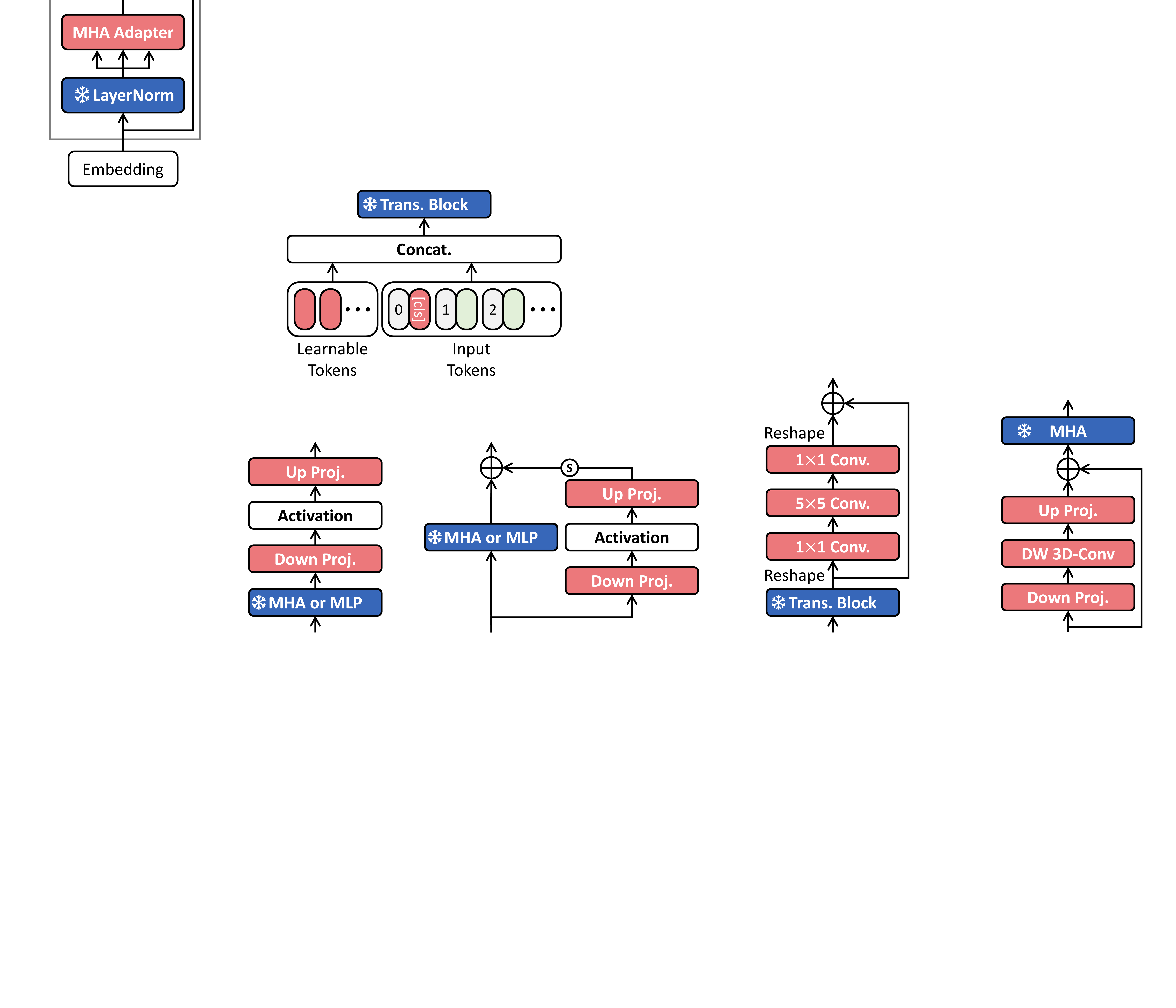}}
            \caption{Serial Adapter~\cite{adaptformer}}\label{fig:2c}
           \end{subfigure}
           \begin{subfigure}[t]{0.32\linewidth}
            \centering
            {\includegraphics[width=0.7\linewidth]{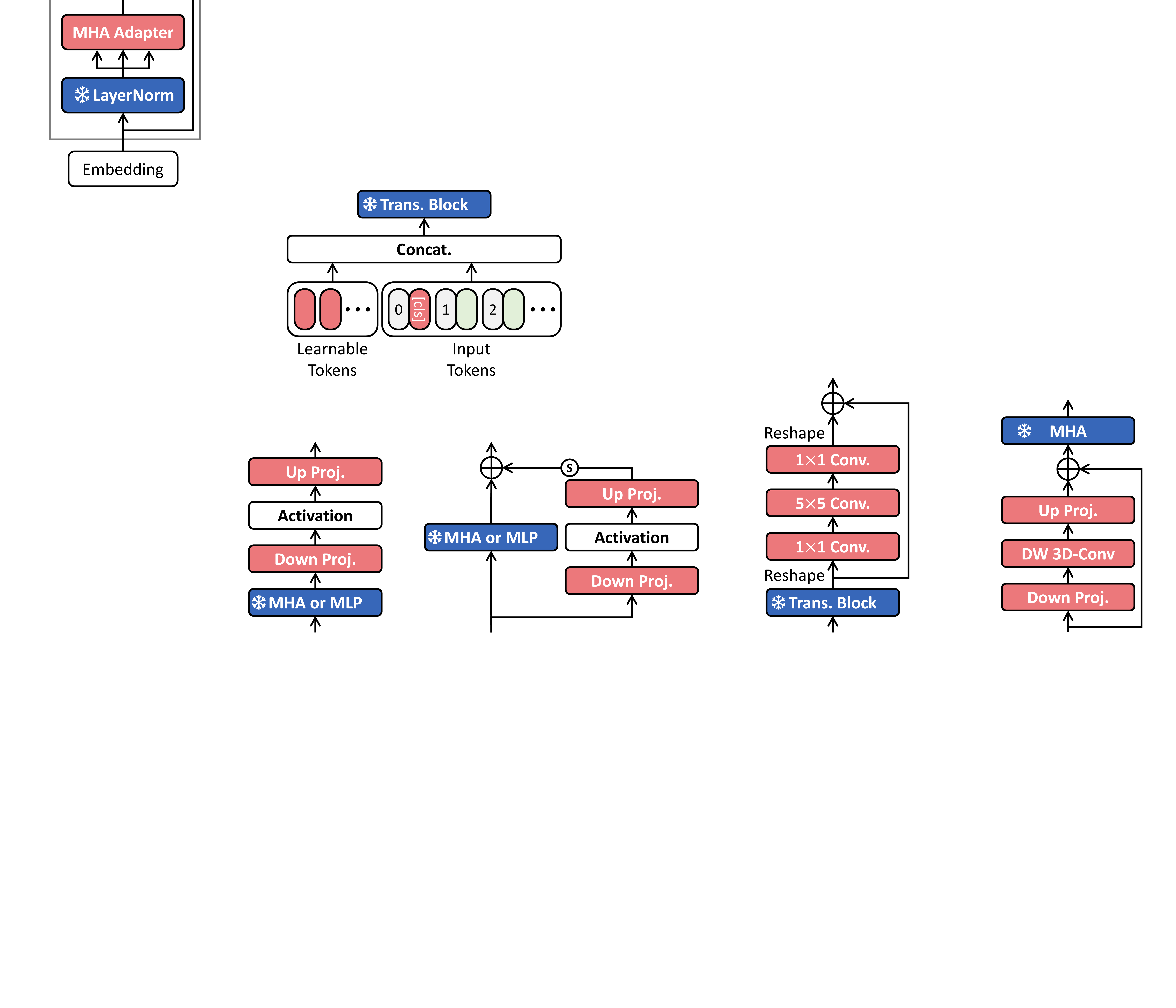}}
            \caption{Prompt Block~\cite{protuning}}\label{fig:2d}
           \end{subfigure}
           \begin{subfigure}[t]{0.32\linewidth}
            \centering
            {\includegraphics[width=0.7\linewidth]{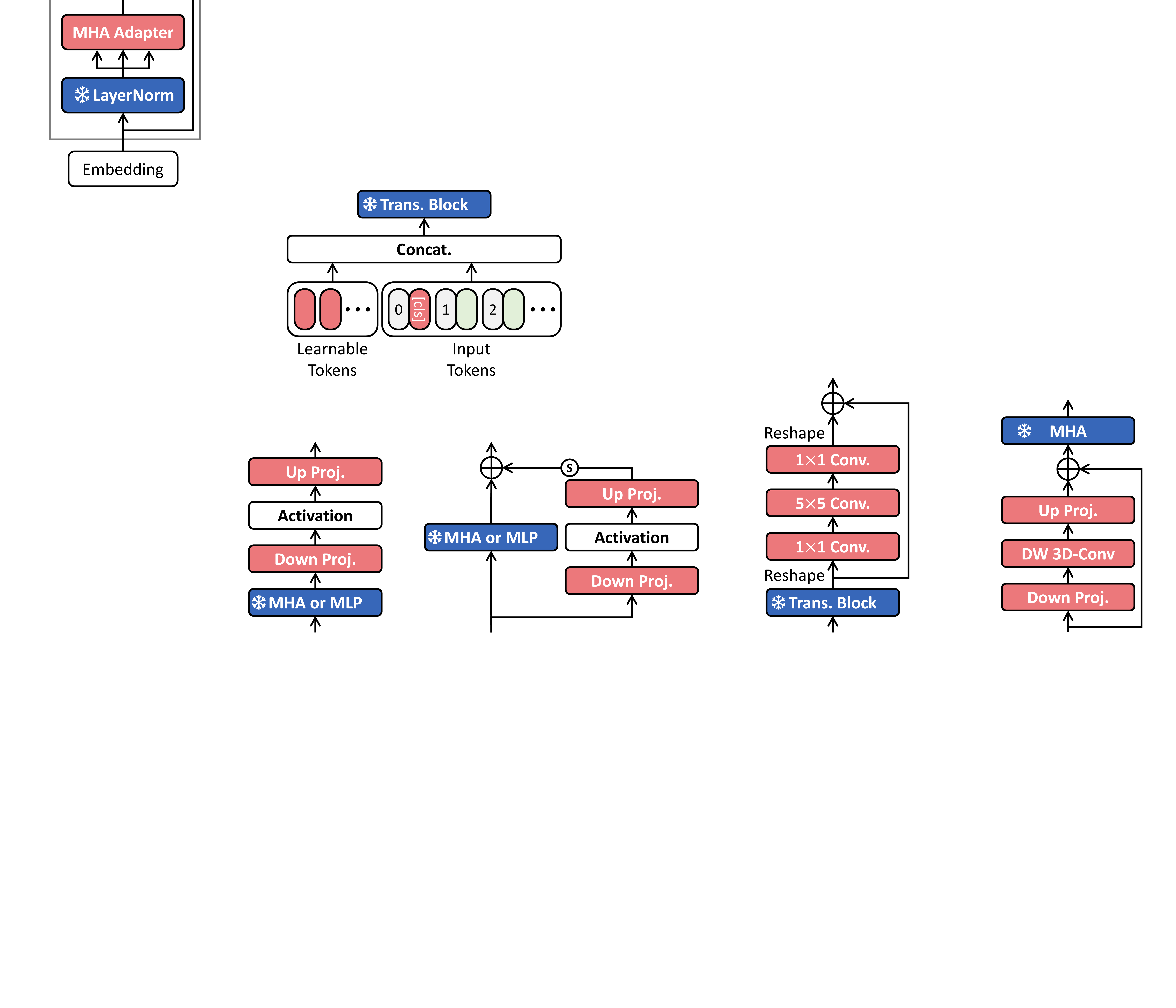}}
            \caption{ST-Adapter~\cite{st-adapter}}\label{fig:2e}
           \end{subfigure}\vspace{-7pt}
        \caption{Baselines for image-to-video transfer learning. (a) Visual Prompt Tuning~\cite{vpt}; (b) Parallel adapter and (c) serial adapter~\cite{adaptformer};
        (d) Pro-Tuning~\cite{protuning}; and (e) ST-Adapter~\cite{st-adapter}.}
        \vspace{-5pt}
    \label{fig:2}
    \end{figure}

    \subsection{Baselines}\vspace{-5pt}
    The objective of our work is to transfer the superiority of vision transformers pretrained on large-scale image datasets to the video domain through efficient finetuning with a small number of learnable parameters, while freezing the pretrained parameters.
    To compare with other methods, we generalize four recent PETL methods to the video domain only with the least possible transformation; (1) VPT~\cite{vpt} (2) AdaptFormer~\cite{adaptformer} (3) Pro-tuning~\cite{protuning} (4) ST-adapter~\cite{st-adapter}.
    The most of works have been originally proposed to adapt pretrained image models to downstream image tasks~\cite{vpt, adaptformer, protuning} and video models to video tasks~\cite{adaptformer}, by learning visual prompt tokens~\cite{vpt}, adapter blocks~\cite{adaptformer} and prompt prediction blocks~\cite{protuning}.
    Only the ST-adapter~\cite{st-adapter} has proposed image-to-video transfer learning.
    In this section, we describe baselines for image-to-video transfer learning in detail.
    For brevity, we leave out the subscripts in \Eq{transformer} and arouse them as needed.
    In addition, we represent the \learn{\textbf{learnable}} and \frozen{\textbf{frozen}} parameters in red and blue colors, respectively.

    \begin{figure*}[t]
    \centering
        \includegraphics[width=1.0 \linewidth]{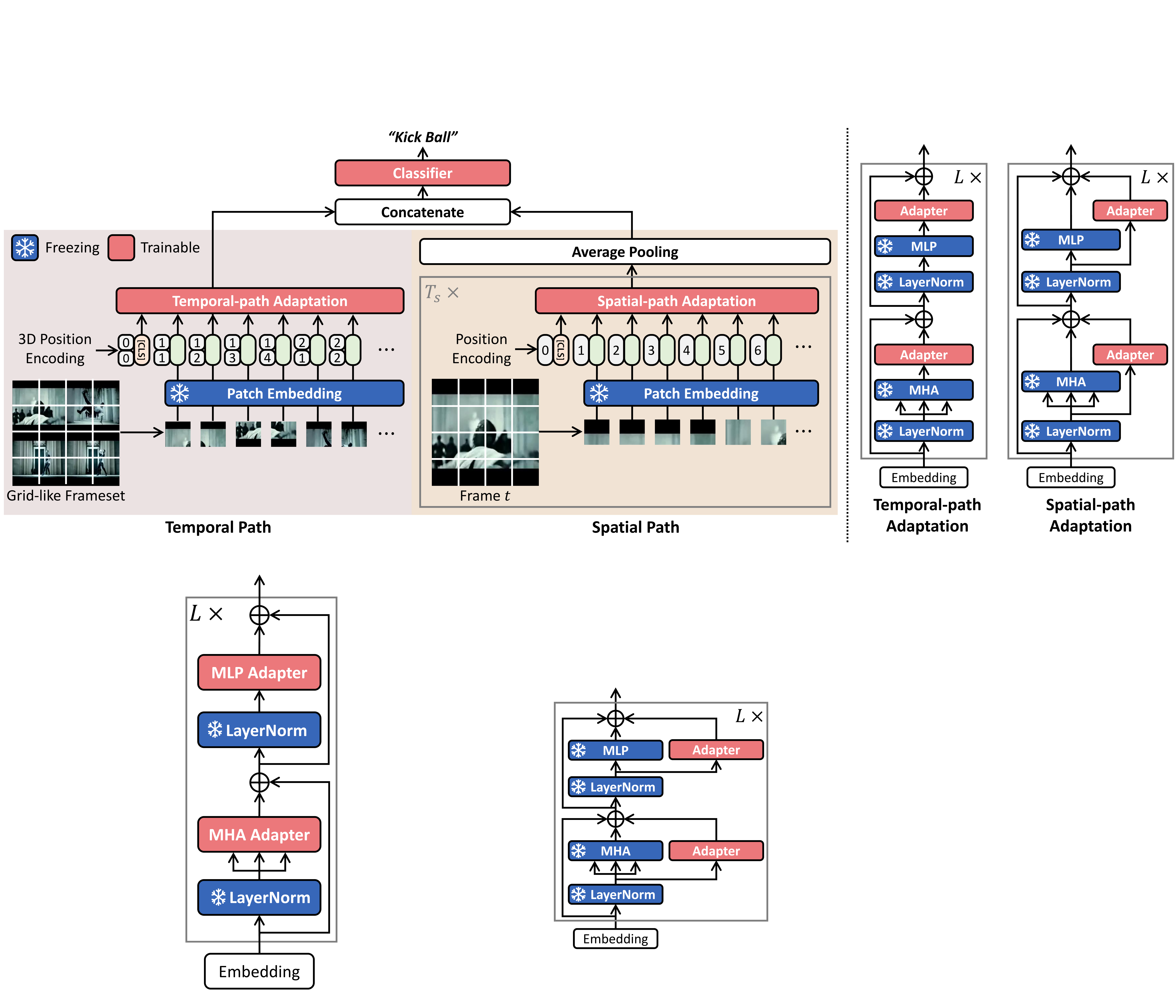} \\ \vspace{-10pt}
      \caption{Overall architecture of \method. The pretrained image transformer takes a grid-like frameset and $T_s$ frames as inputs. We learn the temporal and spatial contexts separately through two distinct paths.
      During training, we only update newly introduced adapters and the classifier while freezing the pretrained layers.}\vspace{-5pt}
    \label{fig:main}
\end{figure*}

    \noindent\textbf{Visual prompt tuning (VPT)}~\cite{vpt}
    prepends $K$ trainable prompt tokens to the input space of every transformer block\footnote{While the original work also presented a shallow version (VPT-Shallow) that inserts prompt tokens to the first layer, we explore a deep version (VPT-Deep) only.} while keeping pretrained parameters frozen.
    The input tokens for each transformer block can be written as:
    \begin{equation}
        \tilde{\bh} = [\learn{\be} ;\bh],
    \end{equation}
    where $\learn{\be} \in \mathbb{R}^{K \times d}$ is a set of trainable visual prompt tokens and $d$ is a channel dimension of the original token.

    \noindent\textbf{AdaptFormer}~\cite{adaptformer}
    learns a trainable bottleneck module~\cite{adapter-nlp1, adapter-nlp2}. 
    The intermediate feature $\bz$ in \Eq{transformer} of each transformer block is fed into the AdapterMLP that consists of the original MLP layers and a bottleneck structure (parallel adapter in \figref{fig:2b}).
    The output of the AdaptFormer block can be formulated by:
    \begin{equation}
    \begin{aligned}
        % \tilde{\mathbf{h}}' &= \frozen{\texttt{LN}}(\mathbf{h}'),    \\
        \tilde{\bz} & = \sigma(\frozen{\lnorm}(\bz)\cdot \learn{\mathbf{W}_{\text{down}}}) \cdot \learn{\mathbf{W}_{\text{up}}},   \\
        \mathbf{h}& = \bz + \frozen{\mlp}(\frozen{\lnorm}(\bz)) + s \cdot \tilde{\bz},
    \end{aligned}\label{equ:adapter}
    \end{equation}
    where $\learn{\mathbf{W}_{\text{down}}}, \learn{\mathbf{W}_{\text{up}}}$ are trainable down- and up-projection matrices, $\sigma(\cdot)$ is an activation function, and $s$ is a scaling factor.

    \noindent\textbf{Pro-tuning}~\cite{protuning}
    predicts task-specific vision prompts $\bv$ from the output of each transformer block using consecutive 2D convolution layers.
    The output of each block is reshaped as $\mathbb{R}^{P\times P \times C}$ to apply 2D convolutions and the final representation is derived by adding $\bv$ and $\bh$:
    \begin{equation}
    \begin{aligned}
        \bv = \reshape(\sigma(&\learn{\conv}(\reshape(\bh)))),  \\        
        \tilde{\bh} &= \bh + \bv,   \\
    \end{aligned}
    \end{equation}
    where $\learn{\texttt{Conv2D}}$ consists of $1\times 1$ convolution layer followed by $5\times 5$ depth-wise convolution~\cite{dw-convolution} and $1\times 1$ convolution.

    \noindent\textbf{ST-adapter}~\cite{st-adapter}
    inserts a depth-wise 3D convolution layer between the down-projection layer and the activation function of the adapter.
    Different from the conventional adapters (\eg AdaptFormer~\cite{adaptformer}), the ST-adapter takes tokens for all frames to enable the model to capture temporality in videos.
    The output of the ST-adapter can be represented as:
    \begin{equation}
    \begin{aligned}
        \tilde{\bz}_t &= \frozen{\lnorm}(\bz_t),    \\
        \hat{\bz} = \sigma(\learn{\dwconv}([&\tilde{\bz}_1{\cdot} \learn{\mathbf{W}_{\text{down}}}, \cdots, \tilde{\bz}_T{\cdot} \learn{\mathbf{W}_{\text{down}}}])){\cdot}\learn{\mathbf{W}_{\text{up}}},  \\
        \bh_t = \bz_t +& \frozen{\mlp}(\tilde{\bz}_t) + s \cdot \hat{\bz}_t,        
    \end{aligned}
    \end{equation}
    where $\learn{\dwconv}$ denotes the depth-wise 3D convolution layer.
    Note that the same down-projection matrix $\learn{\mathbf{W}_{\text{down}}}$ is applied to all tokens regardless of the frame index $t$.

    We emphasize that most of the baselines~\cite{vpt, adaptformer, protuning} have \textit{not} concerned with temporal modeling. Even though ST-Adapter~\cite{st-adapter} has employed depth-wise 3D convolution layers between linear projections, it results in high computational cost. To entirely leverage a simple and efficient architecture of the adapter~\cite{adaptformer}, we incorporate the dual-path design into the pretrained image transformers.

\section{Method}  
\vspace{-5pt}
    The dual-path design (also called two-stream) is well-known architecture in CNN-based models for video recognition~\cite{i3d, two-stream, slowfast}.
    They have commonly used an optical flow~\cite{i3d, two-stream} or multiple frames with a high temporal resolution~\cite{slowfast} to capture rapidly changing motion.
    Despite the effectiveness of dual-path architecture, it has yet to be explored with the transformer due to high computational costs.
    In this work, we propose a novel PETL method, called \method, comprised of \textit{spatial} and \textit{temporal} path adaptation.
    To the best of our knowledge, our \method is the first attempt to explicitly build the two-stream architecture upon the transformer while maintaining the computational cost similar to the single-stream architecture.
    The overall framework is depicted in \figref{fig:main}.

    \vspace{-5pt}
    \subsection{Spatial adaptation}\label{sec:sa}\vspace{-5pt}
    Since the image foundation models have been trained on large amounts of web datasets, we can intuitively speculate that they might be powerful to encode the spatial context even in videos.
    In order to make the outstanding ability of spatial modeling to be more suitable for video understanding with a slight parameter tuning, we adopt two parallel adapters for $\mha$ and $\mlp$ in each transformer block.
    The parallel adapters allow the model to learn the spatial context for action recognition from the appearance of target video data while maintaining the original contexts for object recognition.
    
    Specifically, we sample $T_S$ frames from a video and tokenize each frame.
    Similar to \Eq{tokenization},
    the set of spatial tokens $\mathbf{X}^{\text{SP}}_t$ includes the learnable positional encodings $\learn{\bpsp}$ and the spatial class token $\learn{\spcls}$.
    The spatial adaptation in the $l$-th transformer block can be formulated by the following equations:
    \begin{equation}
    \small
        \begin{aligned}
            \bz_{l}^{\text{SP}} = \bh_{l-1}^{\text{SP}} &+ \frozen{\mha}(\frozen{\lnorm}(\bh_{l-1}^{\text{SP}})) + \learn{\texttt{Adapter}}(\frozen{\lnorm}(\bh_{l-1}^{\text{SP}})), \\
            \bh_{l}^{\text{SP}} = \bz_{l}^{\text{SP}} &+ \frozen{\mlp}(\frozen{\lnorm}(\bz_{l}^{\text{SP}})) + \learn{\texttt{Adapter}}(\frozen{\lnorm}(\bz_{l}^{\text{SP}})),
        \end{aligned}
    \end{equation}
    where $\mathbf{h}_{0}^{\text{SP}} = [\learn{\spcls}, \mathbf{X}_t^{\text{SP}}]+\learn{\bpsp}$.
    We average the set of the spatial $\cls$ tokens from the final transformer block to obtain a global spatial representation $\mathbf{y}^{\text{SP}}$, such that,
    \begin{equation}
        \mathbf{y}^{\text{SP}} = \frac{1}{T_S}\sum_{t=1}^{T_S} \spcls.
    \end{equation}
    Recent methods have discussed that a high frame rate only increases the computation volume and is unnecessary to understand the semantics of appearance~\cite{slowfast, revisiting}.
    We thus sparsely sample $T_S$ frames with a low frame rate (\eg 8 frames per clip).

    \subsection{Temporal adaptation}\label{sec:ta}\vspace{-5pt}
    While spatial adaptation enables the models to take the spatial contexts in video data, the image models are still incapable of modeling the temporal dynamics.
    The key component that allows video transformers to model the solid temporal context is to learn relationships between local patches across \textit{frames} in the video~\cite{space-time, vivit}.
    To make image models capable of effectively establishing this component, we suggest a novel \textit{grid-like frameset} transform technique that aggregates multiple frames into a unified \textit{grid-like frameset}.
    Our grid-like frameset design is inspired by recent visual prompting research~\cite{inpainting, prompting}.
    It is simple yet surprisingly effective in imitating temporal modeling as spatial modeling and certainly reduces the computation.
    In each transformer block, we adopt two additional serial adapters for $\mha$ and $\mlp$, respectively.

    More concretely, we sample $T$ frames from a video and scale them with factors of $w$ and $h$, such that the scaled frame size is $[W/w \times H/h \times 3]$.
    We stack $w\times h$ scaled frames according to temporal ordering and reshape a stacked frame to construct a set of frames in a grid form of the same size as the original frame (\ie, $[W \times H \times 3]$).
    Note that the total number of grid-like framesets is $T_G = T/wh$.
    The set of temporal tokens $\mathbf{X}_g^{\text{TP}}$ for the $g$-th frameset is obtained in the same way in \Eq{tokenization} and combined with the learnable temporal class token $\learn{\tpcls}$.
    Unlike the spatial adaptation, we use fixed 3D positional encodings~\cite{fixed-position}, $\bptp$, to the tokens to take the absolute temporal order and spatial positions of patches into account.
    The input transformation allows transformers to observe multiple frames at the same level.
    In experiments, we mainly construct a grid-like frameset from 16 original frames (\ie, scaling factors $w=h=4$) to take the computational efficiency and promising performance.
    
    Whereas the parallel adapter is used in the spatial path, we sequentially append adapters to the top of \texttt{MHA} and \texttt{MLP} layers in each transformer block.
    Formally, temporal adaptation in the $l$-th block can be described as:
    \begin{equation}
        \begin{aligned}
            \bz_{l}^{\text{TP}} = \mathbf{h}_{l-1}^{\text{TP}} &+ \learn{\texttt{Adapter}}(\frozen{\texttt{MHA}}(\frozen{\texttt{LN}}(\mathbf{h}_{l-1}^{\text{TP}}))), \\
            \mathbf{h}_{l}^{\text{TP}} = \bz_{l}^{\text{TP}} &+ \learn{\texttt{Adapter}}(\frozen{\texttt{MLP}}(\frozen{\texttt{LN}}(\bz_{l}^{\text{TP}}))),
        \end{aligned}
    \end{equation}
    where $\bh^{\text{TP}}_{0} = [\learn{\tpcls}, \mathbf{X}_g^{\text{TP}}] + \bptp$.
    Similar to spatial adaptation, a global temporal representation $\by^{\text{TP}}$ can be derived by averaging the set of the temporal $\cls$ tokens from the final transformer block, \ie,
    \begin{equation}
        \by^{\text{TP}} = \frac{1}{T_G} \sum_{g=1}^{T_G} \mathbf{x}_g^{\text{TP}}\{\cls\}.
    \end{equation}

    For the final prediction, we concatenate the global spatial and temporal representations and feed them into the classifier with GeLU activation~\cite{gelu} between two FC layers.
    
     \begin{figure}[t]
        \centering
            \begin{subfigure}{0.48\linewidth}
            \centering
            {\includegraphics[width=0.95\linewidth]{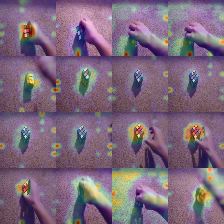}}
            \caption{\method w/o TA}\label{fig:quala}
           \end{subfigure}
           \begin{subfigure}{0.48\linewidth}
            \centering
            {\includegraphics[width=0.95\linewidth]{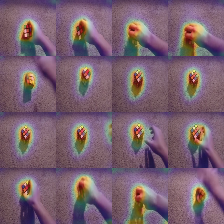}}
            \caption{\method w/ TA}\label{fig:qualb}
           \end{subfigure}\vspace{-7pt}
        \caption{Visualization of attention maps (a) without temporal adaptation (TA) and (b) with temporal adaptation for the action 'Spinning [something] that quickly stops spinning' in SSv2~\cite{ssv2}.}\vspace{-5pt}
    \label{fig:qual}
    \end{figure}
    \begin{table*}[t]
    \centering
    \small
    \setlength{\tabcolsep}{7pt}
        \begin{tabular}{l c c c c c c c c }
        \toprule
        Method \& Arch. &   Pretrain    &  \makecell{Model \\ \# Params}   & \makecell{Trainable \\ \# Params}      & GFLOPs    & R@1 & R@5 &   Views  \\ 
        \midrule
        \multicolumn{5}{l}{\hspace{-6pt}\textit{Full-tuning}} \\ 
        SlowFast+NL~\cite{slowfast}          &    -   &   60M     &  60M  &     7020     &  79.8  & 93.9    &   16$\times$3$\times$10   \\
        MViT-B~\cite{multiscale}          &    -   &   37M     &  37M  &     4095     &  81.2  & 95.1    &   64$\times$3$\times$3   \\
        UniFormer-B~\cite{uniformer}          &    IN-1K   &   50M     &  50M  &     3108     &  83.0  & 95.4    &   32$\times$4$\times$3   \\
        TimeSformer-L~\cite{space-time}          &    IN-21K   &   121M     &  121M  &     7140     &  80.7  & 94.7    &   64$\times$1$\times$3   \\
        ViViT-L/16$\times$2 FE~\cite{vivit}          &    IN-1K   &   311M     &  311M  &     3980     &  80.6  & 92.7    &   32$\times$1$\times$1   \\
        VideoSwin-L~\cite{video-swin}          &    IN-21K   &   197M     &  197M  &     7248     &  83.1  & 95.9    &   32$\times$4$\times$3   \\
        MViTv2-L~\cite{multiscale-v2}          &    IN-21K   &   218M     &  218M  &     42420     &  86.1  & 97.0    &   32$\times$3$\times$5   \\
        MTV-L~\cite{multiview}         &    JFT   &   876M     &  876M  &     18050     &  84.3  & 96.3    &   32$\times$4$\times$3   \\
        TokenLearner-L/10~\cite{tokenlearner}         &    JFT   &   450M     &  450M  &     48912     &  85.4  & 96.3    &   64$\times$4$\times$3   \\
        ActionCLIP~\cite{actionclip}         &    CLIP   &   142M     &  142M  &     16890     &  83.8  & 97.1    &   32$\times$10$\times$3   \\
        X-CLIP-L/14~\cite{x-clip}          &    CLIP   &   420M     &  420M  &     7890     &  87.1  & 97.6    &   8$\times$4$\times$3   \\ 
        
        \midrule
        \multicolumn{5}{l}{\hspace{-6pt}\textit{Parameter Efficient Tuning}} \\ 
        EVL~\cite{frozen-clip} w/ ViT-L/14          &    CLIP   &   368M     &  59M  &     8088     &  87.3  & -    &   32$\times$3$\times$1   \\ 
        ST-Adapter~\cite{st-adapter} w/ ViT-B/16     &    CLIP   &   93M     &  7M  &    1821      &  82.7  & 96.2    &   32$\times$3$\times$1   \\
        \rowcolor{Light} \textbf{\method} w/ ViT-B/16        &  CLIP  &  96M  &  10M  &  710 &85.4  &97.1   &  32$\times$3$\times$1  \\
        \rowcolor{Light} \textbf{\method} w/ ViT-L/14        &  CLIP  &  330M  &  27M  &  1868 & 
 87.7  &  97.8   &  32$\times$3$\times$1  \\
        \bottomrule
        \end{tabular}\vspace{-5pt}
    \caption{Performance comparisons for action recognition on the Kinetics-400~\cite{k400} dataset. 
    Note that Views = \#frames $\times$ \#clips $\times$ \#spatial.}\vspace{-5pt}\label{tab:k400}
    \end{table*}

    \subsection{Does a grid-like frameset really help to encode temporal context?}\vspace{-5pt}
    This work presents a new standpoint to perform video action recognition with the pretrained image transformer by transforming multiple frames into a unified grid-like frameset.
    However, it is still questionable whether the temporal path of \method can really capture the temporal context of videos.
    In this section, we provide some qualitative examples of the attention map.
    To validate the ability of precise temporal modeling, we sample videos from the SSv2~\cite{ssv2} dataset.
    Following \cite{st-adapter}, we depict the attention map of $\tpcls$ from the final transformer block of the temporal path.
    As shown in \figref{fig:qual}, the model with the temporal adaptation (TA) of \method tends to concentrate on action-related regions, contrary to the model without TA that focuses on the irrelevant background.
    This result suggests that the temporal adaptation of \method strengthens the temporal modeling ability of the image model.
    More examples are shown in \figref{fig:qual_supp} of Appendix.
\section{Experiment}
\vspace{-3pt}
\subsection{Evaluation setup} \vspace{-5pt}
\paragrapht{Datasets.} 
    We evaluate the proposed method on four standard action recognition datasets, including the Kinetics-400 (K400)~\cite{k400}, HMDB51~\cite{hmdb51}, Something-something-v2 (SSv2)~\cite{ssv2}, and Diving-48~\cite{diving}. \vspace{-5pt}
\begin{itemize}
    \item{\textbf{Kinetics-400 (K400)} contains about 240K training videos and 20K validation videos for 400 human action categories.
    Each video is trimmed to have a length of 10 seconds.
    While the K400 dataset provides a wide range of categories, they are known to be highly biased in spatial appearance~\cite{k400-bias}.}\vspace{-5pt}
    \item{\textbf{Somthing-something-v2 (SSv2)} is a more challenging dataset since they require strong temporal modeling~\cite{k400-bias}.
    They contain about 168.9K training videos and 24.7K validation videos for 174 classes.}\vspace{-5pt}
    \item{\textbf{HMDB51} is a small dataset that provides about 7K videos of 51 action categories.
    We use all three splits, each split of which consists of 3570 and 1530 videos for training and evaluation, respectively.
    We report the average accuracy for three splits.}\vspace{-5pt}
    \item{\textbf{Diving-48} is a fine-grained diving action dataset.
    We use train-test split v2 that contains about 15K training videos and 2K validation videos of 48 diving actions.
    Since the action can not be determined by only the static representations (\eg objects or background), stronger temporal modeling is required for this dataset.}\vspace{-5pt}
\end{itemize}
    
        \begin{table*}[t]
    \centering
    \small
    \setlength{\tabcolsep}{7pt}
        \begin{tabular}{lccccccc}
        \toprule
        Method \& Arch.   &   Pretrain    &  \makecell{Model \\ \#Params}   & \makecell{Trainable \\ \#Params}      & GFLOPs    & R@1 & R@5 &   Views  \\ 
        \midrule
        \multicolumn{5}{l}{\hspace{-5pt}\textit{Full-tuning}} \\
        \rowcolor{light_gray}
        Full-tuning ViT-B/16~\cite{vit}    &    CLIP    &       86M  &   86M  &   419 &   44.0   &   77.0   & 8$\times$1$\times$3    \\
        \rowcolor{light_gray}
        Full-tuning ViT-L/14~\cite{vit}    &    CLIP    &       303M  &   303M  &   1941 &   48.7   &   77.5   & 8$\times$1$\times$3    \\ 
        TimeSformer-L~\cite{space-time}           &    IN-21K   &   121M     &  121M  &     7140     &  62.4  & -    &   64$\times$1$\times$3   \\
        MTV-B~\cite{multiview}           &    IN-21K   &   310M     &  310M  &     4790     &  67.6  & 90.4    &   32$\times$4$\times$3   \\
        MViT-B~\cite{multiscale}           &    K400   &   37M     &  37M  &     510     &  67.1  & 90.8    &   32$\times$1$\times$3   \\
        MViTv2-B~\cite{multiscale-v2}           &    K400   &   51M     &  51M  &     675     &  70.5  & 92.7    &   40$\times$1$\times$3   \\
        ViViT-L/16$\times$2~\cite{vivit}           &    IN-21K/K400   &   311M     &  311M  &     11892     &  65.4  & 89.8    &   16$\times$4$\times$3 \\
        VideoSwin-B~\cite{video-swin}           &    IN-21K/K400   &   89M     &  89M  &     963     &  69.6  & 92.7    &   32$\times$1$\times$1   \\
        Omnivore~\cite{omnivore}           &    IN-21K/K400   &   -     &  -  &     -     &  71.4  & 93.5    &   32$\times$1$\times$3   \\
        MViTv2-B~\cite{multiscale-v2}           &    IN-21K/K400   &   213M     &  213M  &     8484     &  73.3  & 94.1    &   32$\times$1$\times$3   \\
        UniFormer-B~\cite{uniformer}           &    IN-21K/K600   &   50M     &  50M  &     777     &  71.2  & 92.8    &   32$\times$1$\times$3   \\ 
        \midrule
        \multicolumn{5}{l}{\hspace{-5pt}\textit{Parameter Efficient Tuning}} \\ 
        VideoPrompt$^*$~\cite{videoprompt} w/ ViT-B/16  &    CLIP    &       92M  &   6M  &   537 &   31.0   &   60.3   & 8$\times$1$\times$3    \\
        VPT~\cite{vpt} w/ ViT-B/16 &    CLIP    &       92M  &   6M  &   537 &   36.2   &   61.1   & 8$\times$1$\times$3    \\
        AdaptFormer~\cite{adaptformer} w/ ViT-B/16 &    CLIP    &       94M  &   8M  &   544 &   51.3   &   70.6   & 8$\times$1$\times$3    \\
        Pro-tuning~\cite{protuning} w/ ViT-B/16    &    CLIP    &       95M  &   9M  &   538 &   50.8   &   69.9   & 8$\times$1$\times$3    \\
        EVL~\cite{frozen-clip} w/ ViT-L/14  &    CLIP   &   484M     &  175M  &     9641     &  66.7  & -    &   32$\times$1$\times$3   \\ 
        ST-Adapter~\cite{st-adapter} w/ ViT-B/16           &    CLIP   &   97M     &  11M  &    1955      &  69.5  &  92.6    &   32$\times$3$\times$1   \\
        \rowcolor{Light}
        \textbf{\method} w/ ViT-B/16 &  CLIP  &  99M  &  13M  &   642    &   69.6  & 92.5   &  16$\times$1$\times$3  \\
        \rowcolor{Light}
        \textbf{\method} w/ ViT-B/16 &  CLIP  &  99M  &  13M  &   716    & 70.3  &   92.9   &  32$\times$1$\times$3  \\
        \rowcolor{Light}
        \textbf{\method} w/ ViT-B/16 &  CLIP  &  99M  &  13M  &   791    &   71.2  & 93.2   &  48$\times$1$\times$3  \\
        \rowcolor{Light}
        \textbf{\method} w/ ViT-L/14 &  CLIP  &  336M  &  33M  &  1713  & 70.2  &  92.7   &  16$\times$1$\times$3  \\
        \rowcolor{Light}
        \textbf{\method} w/ ViT-L/14 &  CLIP  &  336M  &  33M  &  1932 &   71.4    &   93.4   &  32$\times$1$\times$3  \\
        \rowcolor{Light}
        \textbf{\method} w/ ViT-L/14 &  CLIP  &  336M  &  33M  &  2151 &  72.2  &  93.7   &  48$\times$1$\times$3  \\
        \bottomrule
        \end{tabular}\vspace{-5pt}
    \caption{Performance comparisons for action recognition on the SSv2~\cite{ssv2} dataset. Note that we reproduce VideoPrompt~\cite{videoprompt} without the additional text branch for fair comparison (denoting with $*$).
    }\vspace{-5pt}\label{tab:ssv2}
    \end{table*}

\paragrapht{Pretrained image backbone.}
    We employ CLIP pretrained ViT-B/16 and ViT-L/14 as backbones. The results with Swin-B~\cite{swin} are provided in \tabref{tab:add_results} of the Appendix.\vspace{-5pt}
    \begin{itemize}
        \item{\textbf{ViT-B/16}}~\cite{vit} consists of 12 transformer blocks with 86M parameters and takes patches of size $16\times 16$ as inputs.
        \vspace{-5pt}
        \item{\textbf{ViT-L/14}}~\cite{vit}, a larger model than ViT-B/16, contains 24 transformer blocks with 303M parameters. It takes $14\times 14$ patches as inputs.
        \vspace{-5pt}
    \end{itemize}

\paragrapht{Implementation details.}
    For the K400, HMDB51, and Diving-48 datasets, we uniformly sample 8 frames (\ie, $T_s = 8$) with the sampling interval 8 in the spatial path.
    In the temporal path, we uniformly sample 16, 32, and 48 frames with the sampling intervals 4, 2, and 1 to construct 1, 2, and 3 grid-like framesets (\ie, $T_g = 1, 2, 3$), respectively.
    For the SSv2 dataset, we sample the same number of frames as in other datasets, but with a dynamic sampling interval to cover the whole video.
    Note that the frames for the spatial path are the subset of the temporal path frames.
    Please refer to more implementation details in Appendix~\ref{sec:imple_detail}.

\subsection{Comparison with state-of-the-art} \vspace{-5pt}
    In this section, we compare the proposed method with baselines~\cite{adaptformer, vpt, protuning, st-adapter} in \secref{sec:baseline} and state-of-the-art video transformers~\cite{multiscale, uniformer, space-time, vivit, video-swin, multiscale-v2, multiview, tokenlearner, actionclip, x-clip, frozen-clip} to demonstrate the effectiveness of the proposed method on four video action recognition datasets.
    Note that the number of frames of the spatial adaptation path of \method is fixed to 8 for all experiments, \ie, $T_S=8$.
    
    \paragrapht{Results on Kinetics-400.}
    We report the results evaluated on K400~\cite{k400} in \tabref{tab:k400}.
    We first compare the proposed method with state-of-the-art video models that are pretrained on the large-scale image dataset and fully finetuned on K400.
    In terms of memory and computational efficiency, video models require a huge number of parameters ($\sim$450M~\cite{tokenlearner}) and computations ($\sim$48912 GFLOPs~\cite{tokenlearner}).
    On the other hand, we require only 10M trainable parameters which are newly stored, and 710 GFLOPs for inference using 32 frames with ViT-B/16~\cite{vit} backbone.
    Compared to X-CLIP-L/14~\cite{x-clip} which leverages the additional text branch, our \method achieves state-of-the-art performance with ViT-L/14 backbone.
    The comparisons with parameter-efficient tuning methods~\cite{frozen-clip, st-adapter} show that our \method achieves higher performance while requiring much lower burdens in computations under the same conditions.
    
    \paragrapht{Results on Something-Something-v2.}
    We present the performance comparisons on SSv2~\cite{ssv2} in \tabref{tab:ssv2}.
    The results show that our \method with ViT-B/16 achieves comparable or better performance than the prior supervised video models~\cite{space-time, multiview, multiscale, vivit}, requiring a much smaller number of trainable parameters and GFLOPs.
    Our \method with ViT-L/14 shows more competitive performance, outperforming most prior works~\cite{multiscale-v2, uniformer, omnivore}.
    The baselines~\cite{videoprompt, vpt, adaptformer, protuning}, which have relatively weak temporal modeling ability, show significantly poor performance, implying that strong temporal modeling is mandatory for the SSv2 dataset.
    The comparisons to the CLIP pretrained PET approaches~\cite{frozen-clip, st-adapter} with ViT-B/16 demonstrate the effectiveness and efficiency of \method, achieving higher performance (70.3 vs 69.5~\cite{st-adapter}) with significantly low computations (716 vs 9641~\cite{frozen-clip} GFLOPs) using 32 frames.
    Thanks to the extreme computational efficiency, our \method comprises more competitive performance using 48 frames ($T_G=3$) with low computation requirements.
    
    \begin{table}[t]
    \centering
    \small
    \setlength{\tabcolsep}{4pt}
        \begin{tabular}{lccc}
        \toprule
        Method \& Arch.   & {Classifier}  & {Params}   & {HMDB51}  \\
        \midrule
        \rowcolor{light_gray}
        Full-tuning w/ ViT-B/16~\cite{vit} & Lin.  &   86M  &   59.4   \\
        Linear w/ ViT-B/16 & Lin.  &  0.1M  &   61.2   \\
        VPT~\cite{vpt} w/ ViT-B/16 &   Trans.   &   7M    &   62.4        \\
        AdaptFormer~\cite{adaptformer} w/ ViT-B/16  &   Trans.    &   8M       &   63.7   \\
        Pro-tuning~\cite{protuning} w/ ViT-B/16 &   Trans. &   9M    &  63.3    \\
        VideoPrompt~\cite{videoprompt} w/ ViT-B/16  &   Trans.   &   6M    &  66.4    \\
        ST-Adapter${^*}$~\cite{st-adapter} w/ ViT-B/16    &   Lin. &   7M       &   65.9   \\
        \rowcolor{Light}
        \textbf{\method} w/ ViT-B/16 &   MLPs.     &    10M    &  75.6         \\
        \bottomrule
        \end{tabular}\vspace{-7pt}
    \caption{Performance comparisons for action recognition on the HMDB51~\cite{hmdb51} dataset with the CLIP pretrained ViT-B/16~\cite{clip}. We report the type of classifier and the number of learnable parameters for baselines and ours. \textbf{Lin.} and \textbf{Trans.} denote the linear classifier and temporal transformer, respectively. Our \method  uses two MLP layers as the classifier. Note that we reproduce ST-Adapter~\cite{st-adapter} for fair comparison in terms of the pretrained dataset (denoting with ${*}$).
    }\vspace{-10pt}\label{tab:hmdb51}
    \end{table}
        
    \begin{table}[t]
    \centering
    \small
    \setlength{\tabcolsep}{4pt}
        \begin{tabular}{lccc}
        \toprule
        {Method \& Arch.}  & {Pretrain}   & {Params}    &  {Diving48} \\
        \midrule
        \multicolumn{4}{l}{\hspace{-5pt}\textit{Supervised}} \\ 
        TimeSformer-L~\cite{space-time} &  IN-21K &   121M    &   81.0     \\
        VideoSwin-B~\cite{video-swin} &  IN-21K &   88M    &   81.9     \\
        SIFAR-B-14~\cite{sifar} &  IN-21K &   87M    &   87.3     \\
        ORViT~\cite{orvit} &  IN-21K &   160M    &   88.0   \\  \midrule
        \multicolumn{4}{l}{\hspace{-5pt}\textit{Parameter Efficient Tuning}} \\ 
        \rowcolor{Light}
        \textbf{\method} w/ ViT-B/16     &  CLIP &    10M   &  88.7     \\
        \bottomrule
        \end{tabular}\vspace{-7pt}
    \caption{Performance comparisons for action recognition on the Diving-48~\cite{diving} dataset. We report the pretrained dataset, the number of learnable parameters (M) for each method, and the accuracy.
    }\vspace{0pt}\label{tab:diving48}
    \end{table}

    \paragrapht{Results on HMDB51.}
    In \tabref{tab:hmdb51}, we compare the results with baselines~\cite{vpt, adaptformer, protuning, videoprompt, st-adapter} on HMDB51~\cite{hmdb51} that dominantly requires strong spatial modeling for action recognition.
    Surprisingly, our \method significantly outperforms baselines by large margins.
    This result demonstrates \method fully capitalizes on the exceptional spatial modeling ability of the pretrained image model for action recognition.
    The comparison with VideoPrompt~\cite{videoprompt} that uses the additional text branch demonstrates the effectiveness of \method, improving 9.2\% performance improvement.
    
    \paragrapht{Results on Diving-48.}
    \tabref{tab:diving48} shows performance comparisons on Diving-48~\cite{diving} that requires fine-grained action recognition.
    Our \method consistently outperforms all video models with only 10M trainable parameters.
    Particularly, we obtain a better performance than ORViT~\cite{orvit} which utilizes the additional tracking model.
    It indicates the utility of \method in fine-grained action recognition and the superiority of temporal modeling of \method.

    \subsection{Components analysis}\label{sec:ablation}\vspace{-5pt}
    
    \paragrapht{Impact of dual-path.}
    In the top panel of \tabref{tab:ablation}, we train the model by ablating each path and evaluate the performance on SSv2.
    Without the temporal path (\method w/o TA), the performance is significantly degraded despite using a larger number of frames ($T_S\,$=$\,16$ vs $8$).
    Without the spatial path (\method w/o SA), we can obtain slightly better performance than the result without the temporal adaptation.
    Since the SSv2 requires strong temporal modeling, we speculate that two ablation studies show comparison results.
    However, it still shows a substantial performance gap compared to the full model of \method, demonstrating the effectiveness of the dual-path design.

    \paragrapht{Frame rates in spatial path.}
    In the middle panel of \tabref{tab:ablation}, we analyze the effect according to the number of frames used in the spatial path.
    The temporal path identically uses 16 frames to construct a grid-like frameset, and frames used in the spatial path are sampled from such 16 frames.
    A large number of $T_S$ slightly improves performance, however, requires more computational costs.
    Considering the performance improvement compared to the computation increase, we mainly set $T_S$ to 8.

    \paragrapht{Number of frames in grid.}
    We scale down original frames with scaling factors $w$ and $h$ to construct grid-like framesets.
    These factors thus determine the number of frames the model observes within one grid-like frameset.
    While a large value of factors increases the temporal resolution, the information of each frame is inevitably reduced.
    For example, the original frame is scaled down to the size of $28\times 28$ with $w\,$=$\,h\,$=$\,8$.
    Meanwhile, a small value of factors retains richer information from the original frame, however, makes the temporal resolution small.
    As shown in the bottom panel of \tabref{tab:ablation}, we attain the best performance with $w\,$=$\,h\,$=$\,4$.    
        
    \begin{table}[t]
    \centering
    \small
    \setlength{\tabcolsep}{4pt}
        \begin{tabular}{lcccc}
        \toprule
        {Method}     & {Params}  &  {GFLOPs}  &  {SSv2}   & {Views}\\
        \midrule
        \multicolumn{5}{l}{\hspace{-5pt}\textit{Effectiveness of Each Path}} \\ 
        \method w/o TA   &  5M  &  1016 & 53.7  &  16$\times$3$\times$1   \\
        \method w/o SA  &  8M  &   134  &   55.1   &    16$\times$3$\times$1    \\
        \midrule
        \multicolumn{4}{l}{\hspace{-5pt}\textit{Effectiveness of $T_S$}}    \\
        $T_S=8$   &  13M  &  642 & 69.3  &  16$\times$3$\times$1   \\
        $T_S=12$   &  13M  &  896 & 69.6  &  16$\times$3$\times$1  \\
        $T_S=16$   &  13M  &  1150 & 69.8  &  16$\times$3$\times$1   \\  \midrule
        \multicolumn{4}{l}{\hspace{-5pt}\textit{Effectiveness of scaling factors}}    \\
        $w$=$h$=$2 \;(T_G$=$16)$   &  13M  &  1752 & 66.4  &  64$\times$3$\times$1   \\
        $w$=$h$=$4 \;(T_G$=$4)$   &  13M  &  864 & 71.8  &  64$\times$3$\times$1  \\
        $w$=$h$=$8 \;(T_G$=$1)$   &  13M  &  642 & 61.5  &  64$\times$3$\times$1   \\  \midrule
        \rowcolor{Light}
        \textbf{\method}      &  13M &    642   &  69.3  &  16$\times$3$\times$1      \\
        \bottomrule
        \end{tabular}\vspace{-7pt}
        % }
    \caption{Performance with respect to variants of the components.
    }\vspace{-5pt}\label{tab:ablation}
    \end{table}
\vspace{-5pt}
\section{Conclusion and Future Work}
    \vspace{-5pt}
    In this paper, we have introduced the novel image-to-video transfer learning approach, \method.
    By incorporating a dual-path design into image transformers, \method adapts image models to the video task (\ie, action recognition) with a small number of trainable parameters.
    The spatial path adaptation strengthens the inherent spatial context modeling of the pretrained image transformers for video data.
    The temporal path adaptation transforms multiple frames into a unified grid-like frameset, enabling the image model to capture relationships between frames.
    We appropriately employ the bottlenecked adapters in each path to adapt the pretrained features to target video data.
    In addition, we present several baselines transforming recent PETL approaches~\cite{vpt, adaptformer, protuning} for image-to-video adaptation.
    Experimental results demonstrated the superiority of the dual-path design and the grid-like frameset prompting, outperforming several baselines and supervised video models.
    
    There are many possible directions for future work, encompassing cross-domain transfer learning.
    While we have explored image-to-video transfer learning, large foundation vision-language models are available.
    It would also be interesting to expand the superior pretrained 2D knowledge to 3D spatial modeling~\cite{p2p}.
    We hope our study will foster research and provide a foundation for cross-domain transfer learning.
    
\paragrapht{Acknowledgement.}
This research was supported by the Yonsei Signature Research Cluster Program of 2022 (2022-22-0002) and the KIST Institutional Program (Project No.2E31051-21-203).
This research was supported by the National Research Foundation of Korea (NRF) grant funded by the Korea government (MSIP) (NRF2021R1A2C2006703).

%-------------------------------------------------------------------------

%%%%%%%%% REFERENCES
{\small
\bibliographystyle{ieee_fullname}
\bibliography{egbib}

\begin{thebibliography}{10}\itemsep=-1pt

\bibitem{vivit}
Anurag Arnab, Mostafa Dehghani, Georg Heigold, Chen Sun, Mario Lu\v{c}i\'{c},
  and Cordelia Schmid.
\newblock Vivit: A video vision transformer.
\newblock In {\em ICCV}, 2021.

\bibitem{layernorm}
Jimmy~Lei Ba, Jamie~Ryan Kiros, and Geoffrey~E. Hinton.
\newblock Layer normalization.
\newblock {\em arXiv preprint: arXiv:1607.06450}, 2016.

\bibitem{visual-prompt}
Hyojin Bahng, Ali Jahanian, Swami Sankaranarayanan, and Phillip Isola.
\newblock Exploring visual prompts for adapting large-scale models.
\newblock {\em arXiv preprint: arXiv:2203.17274}, 2022.

\bibitem{prompting}
Hyojin Bahng, Ali Jahanian, Swami Sankaranarayanan, and Phillip Isola.
\newblock Exploring visual prompts for adapting large-scale models.
\newblock {\em arXiv preprint: arXiv:2203.17274}, 2022.

\bibitem{inpainting}
Amir Bar, Yossi Gandelsman, Trevor Darrell, Amir Globerson, and Alexei~A.
  Efros.
\newblock Visual prompting via image inpainting.
\newblock {\em arXiv preprint: arXiv:2209.00647}, 2022.

\bibitem{bitfit}
Elad Ben-Zaken, Shauli Ravfogel, and Yoav Goldberg.
\newblock Bitfit: Simple parameter-efficient fine-tuning for transformer-based
  masked language-models.
\newblock In {\em ACL}, 2022.

\bibitem{space-time}
Gedas Bertasius, Heng Wang, and Lorenzo Torresani.
\newblock Is space-time attention all you need for video understanding?
\newblock In {\em ICML}, 2021.

\bibitem{gpt-3}
Tom~B. Brown, Benjamin Mann, Nick Ryder, Melanie Subbiah, Jared Kaplan,
  Prafulla Dhariwal, Arvind Neelakantan, Pranav Shyam, Girish Sastry, Amanda
  Askell, Sandhini Agarwal, Ariel Herbert-Voss, Gretchen Krueger, Tom Henighan,
  Rewon Child, Aditya Ramesh, Daniel~M. Ziegler, Jeffrey Wu, Clemens Winter,
  Christopher Hesse, Mark Chen, Eric Sigler, Mateusz Litwin, Scott Gray,
  Benjamin Chess, Jack Clark, Christopher Berner, Sam McCandlish~Alec Radford,
  Ilya Sutskever, and Dario Amodei.
\newblock Language models are few-shot learners.
\newblock In {\em NeurIPS}, 2020.

\bibitem{revisiting}
Shyamal Buch, Crist\'{o}bal Eyzaguirre, Adrien Gaidon, Jiajun Wu, Li Fei-Fei,
  and Juan~Carlos Niebles.
\newblock Revisiting the ``videos" in video-language understanding.
\newblock In {\em CVPR}, 2022.

\bibitem{i3d}
Jo{\~{a}}o Carreira and Andrew Zisserman.
\newblock Quo vadis, action recognition? a new model and the kinetics dataset.
\newblock In {\em CVPR}, 2017.

\bibitem{adaptformer}
Shoufa Chen, Chongjian Ge, Zhan Tong, Jiangliu Wang, Yibing Song, Jue Wang, and
  Ping Luo.
\newblock Adaptformer: Adapting vision transformers for scalable visual
  recognition.
\newblock In {\em NeurIPS}, 2022.

\bibitem{simclr}
Ting Chen, Simon Kornblith, Mohammad Norouzi, and Geoffrey Hinton.
\newblock A simple framework for contrastive learning of visual
  representations.
\newblock In {\em ICML}, 2020.

\bibitem{empirical}
Xinlei Chen, Saining Xie, and Kaiming He.
\newblock An empirical study of training self-supervised vision transformers.
\newblock In {\em ICCV}, 2021.

\bibitem{dw-convolution}
Fran\c{c}ois Chollet.
\newblock Xception: Deep learning with depthwise separable convolutions.
\newblock In {\em CVPR}, 2017.

\bibitem{randaugment}
Ekin~Dogus Cubuk, Barret Zoph, Jon Shlens, and Quoc Le.
\newblock Randaugment: Practical automated data augmentation with a reduced
  search space.
\newblock In {\em NeurIPS}, 2020.

\bibitem{imagenet}
Jia Deng, Wei Dong, Richard Socher, Li-Jia Li, Kai Li, and Li Fei-Fei.
\newblock Imagenet: A large-scale hierarchical image database.
\newblock In {\em CVPR}, 2009.

\bibitem{cswin}
Xiaoyi Dong, Jianmin Bao, Dongdong Chen, Weiming Zhang, Nenghai Yu, Lu Yuan,
  Dong Chen, and Baining Guo.
\newblock Cswin transformer: A general vision transformer backbone with
  cross-shaped windows.
\newblock In {\em CVPR}, 2022.

\bibitem{vit}
Alexey Dosovitskiy, Lucas Beyer, Alexander Kolesnikov, Dirk Weissenborn,
  Xiaohua Zhai, Thomas Unterthiner, Mostafa Dehghani, Matthias Minderer, Georg
  Heigold, Sylvain Gelly, Jakob Uszkoreit, and Neil Houlsby.
\newblock An image is worth 16x16 words: Transformers for image recognition at
  scale.
\newblock In {\em ICML}, 2021.

\bibitem{multiscale}
Haoqi Fan, Bo Xiong, Karttikeya Mangalam, Yanghao Li, Zhicheng Yan, Jitendra
  Malik, and Christoph Feichtenhofer.
\newblock Multiscale vision transformers.
\newblock In {\em ICCV}, 2021.

\bibitem{sifar}
Quanfu Fan, Chun-Fu~(Richard) Chen, and Rameswar Panda.
\newblock Can an image classifier suffice for action recognition?
\newblock In {\em ICLR}, 2022.

\bibitem{mae-st}
Christoph Feichtenhofer, Haoqi Fan, Yanghao Li, and Kaiming He.
\newblock Masked autoencoders as spatiotemporal learners.
\newblock {\em arXiv preprint: arXiv:2205.09113}, 2022.

\bibitem{slowfast}
Christoph Feichtenhofer, Haoqi Fan, Jitendra Malik, and Kaiming He.
\newblock Slowfast networks for video recognition.
\newblock In {\em ICCV}, 2019.

\bibitem{omnivore}
Rohit Girdhar, Mannat Singh, Nikhila Ravi, Laurens van~der Maaten, Armand
  Joulin, and Ishan Misra.
\newblock Omnivore: A single model for many visual modalities.
\newblock In {\em CVPR}, 2022.

\bibitem{ssv2}
Raghav Goyal, Samira~Ebrahimi Kahou, Vincent Michalski, Joanna Materzyńska,
  Susanne Westphal, Heuna Kim, Valentin Haenel, Ingo Fruend, Peter Yianilos,
  Moritz Mueller-Freitag, Florian Hoppe, Christian Thurau, Ingo Bax, and Roland
  Memisevic.
\newblock The ``something something" video database for learning and evaluating
  visual common sense.
\newblock In {\em ICCV}, 2017.

\bibitem{byol}
Jean-Bastien Grill, Florian Strub, Florent Altch\'e, Corentin Tallec, Pierre~H.
  Richemond, Elena Buchatskaya, Carl Doersch, Bernardo~Avila Pires,
  Zhaohan~Daniel Guo, Mohammad~Gheshlaghi Azar, Bilal Piot, Koray Kavukcuoglu,
  R\'emi Munos, and Michal Valko.
\newblock Bootstrap your own latent: A new approach to self-supervised
  learning.
\newblock In {\em NeurIPS}, 2020.

\bibitem{diff}
Demi Guo, Alexander~M. Rush, and Yoon Kim.
\newblock Parameter-efficient transfer learning with diff pruning.
\newblock In {\em ACL}, 2021.

\bibitem{adapter-nlp2}
Junxian He, Chunting Zhou, Xuezhe Ma, Taylor Berg-Kirkpatrick, and Graham
  Neubig.
\newblock Towards a unified view of parameter-efficient transfer learning.
\newblock In {\em ICLR}, 2022.

\bibitem{mae}
Kaiming He, Xinlei Chen, Saining Xie, Yanghao Li, Piotr Doll\'{a}r, and Ross
  Girshick.
\newblock Masked autoencoders are scalable vision learners.
\newblock In {\em CVPR}, 2022.

\bibitem{moco}
Kaining He, Haoqi Fan, Yuxin Wu, Saining Xie, and Ross Girshick.
\newblock Momentum contrast for unsupervised visual representation learning.
\newblock In {\em CVPR}, 2020.

\bibitem{kaiming-init}
Kaiming He, Xiangyu Zhang, Shaoqing Ren, and Jian Sun.
\newblock Delving deep into rectifiers: Surpassing human-level performance on
  imagenet classification.
\newblock In {\em ICCV}, 2015.

\bibitem{gelu}
Dan Hendrycks and Kevin Gimpel.
\newblock Gaussian error linear units (gelus).
\newblock {\em arXiv preprint: arXiv:1606.08415}, 2016.

\bibitem{orvit}
Roei Herzig, Elad Ben-Avraham, Karttikeya Mangalam, Amir Bar, Gal Chechik, Anna
  Rohrbach, Trevor Darrell, and Amir Globerson.
\newblock Object-region video transformers.
\newblock In {\em CVPR}, 2022.

\bibitem{adapter-nlp1}
Neil Houlsby, Andrei Giurgiu, Stanis\l aw Jastrz\c{e}bski, Bruna Morrone,
  Quentin de Laroussilhe, Andrea Gesmundo, Mona Attariyan, and Sylvain Gelly.
\newblock Parameter-efficient transfer learning for nlp.
\newblock In {\em ICML}, 2019.

\bibitem{lora}
Edward~J. Hu, Yelong Shen, Phillip Wallis, Zeyuan Allen-Zhu, Yuanzhi Li, Shean
  Wang, Lu Wang, and Weizhu Chen.
\newblock Lora: Low-rank adaptation of large language models.
\newblock In {\em ICLR}, 2022.

\bibitem{c3d}
Shuiwang Ji, Wei Xu, Ming Yang, and Kai Yu.
\newblock 3d convolutional neural networks for human action recognition.
\newblock In {\em ICML}, 2010.

\bibitem{align}
Chao Jia, Yinfei Yang, Ye Xia, Yi-Ting Chen, Zarana Parekh, Hieu Pham, Quoc~V.
  Le, Yunhsuan Sung, Zhen Li, and Tom Duerig.
\newblock Scaling up visual and vision-language representation learning with
  noisy text supervision.
\newblock In {\em ICML}, 2021.

\bibitem{vpt}
Menglin Jia, Luming Tang, Bor-Chun Chen, Claire Cardie1, Serge Belongie,
  Bharath Hariharan, and Ser-Nam Lim.
\newblock Visual prompt tuning.
\newblock In {\em ECCV}, 2022.

\bibitem{bypass}
Shibo Jie and Zhi-Hong Deng.
\newblock Convolutional bypasses are better vision transformer adapters.
\newblock {\em arXiv preprint. arXiv:2207.07039}, 2022.

\bibitem{videoprompt}
Chen Ju, Tengda Han, Kunhao Zheng, Ya Zhang, and Weidi Xie.
\newblock Prompting visual-language models for efficient video understanding.
\newblock In {\em ECCV}, 2022.

\bibitem{k400}
Will Kay, Joao Carreira, Karen Simonyan, Brian Zhang, Chloe Hillier, Sudheendra
  Vijayanarasimhan, Fabio Viola, Tim Green, Trevor Back, Paul Natsev, Mustafa
  Suleyman, and Andrew Zisserman.
\newblock The kinetics human action video dataset.
\newblock {\em arXiv preprint: arXiv:1705.06950}, 2017.

\bibitem{hmdb51}
Hildegard Kuehne, Hueihan Jhuang, Est\'ibaliz Garrote, Tomaso Poggio, and
  Thomas Serre.
\newblock Hmdb: A large video database for human motion recognition.
\newblock In {\em ICCV}, 2011.

\bibitem{uniformer}
Kunchang Li, YaliWang, Gao Peng, Guanglu Song, Yu Liu, Hongsheng Li, and Yu
  Qiao.
\newblock Uniformer: Unified transformer for efficient spatial-temporal
  representation learning.
\newblock In {\em ICLR}, 2021.

\bibitem{prefix}
Xiang~Lisa Li and Percy Liang.
\newblock Prefix-tuning: Optimizing continuous prompts for generation.
\newblock In {\em ACL}, 2021.

\bibitem{diving}
Yingwei Li, Yi Li, and Nuno Vasconcelos.
\newblock Resound: Towards action recognition without representation bias.
\newblock In {\em ECCV}, 2018.

\bibitem{multiscale-v2}
Yanghao Li, Chao-Yuan Wu, Haoqi Fan, Karttikeya Mangalam, Bo Xiong, Jitendra
  Malik, and Christoph Feichtenhofer.
\newblock Mvitv2: Improved multiscale vision transformers for classification
  and detection.
\newblock In {\em CVPR}, 2022.

\bibitem{tsm}
Ji Lin, Chuang Gan, and Song Han.
\newblock Tsm: Temporal shift module for efficient video understanding.
\newblock In {\em ICCV}, 2019.

\bibitem{frozen-clip}
Ziyi Lin, Shijie Geng, Renrui Zhang, Peng Gao, Gerard de Melo, Xiaogang Wang,
  Jifeng Dai, Yu Qiao, and Hongsheng Li.
\newblock Frozen clip models are efficient video learners.
\newblock In {\em ECCV}, 2022.

\bibitem{swin-v2}
Ze Liu, Han Hu, Yutong Lin, Zhuliang Yao, Zhenda Xie, Yixuan Wei, Jia Ning, Yue
  Cao, Zheng Zhang, Li Dong, Furu Wei, and Baining Guo.
\newblock Swin transformer v2: Scaling up capacity and resolution.
\newblock In {\em CVPR}, 2022.

\bibitem{swin}
Ze Liu, Yutong Lin, Yue Cao, Han Hu, Yixuan Wei, Zheng Zhang, Stephen Lin, and
  Baining Guo.
\newblock Swin transformer: Hierarchical vision transformer using shifted
  windows.
\newblock In {\em ICCV}, 2021.

\bibitem{video-swin}
Ze Liu, Jia Ning, Yue Cao, Yixuan Wei, Zheng Zhang, Stephen Lin, and Han Hu.
\newblock Video swin transformer.
\newblock In {\em CVPR}, 2022.

\bibitem{cosine-sched}
Ilya Loshchilov and Frank Hutter.
\newblock Sgdr: Stochastic gradient descent with warm restarts.
\newblock In {\em ICLR}, 2017.

\bibitem{adamw}
Ilya Loshchilov and Frank Hutter.
\newblock Decoupled weight decay regularization.
\newblock In {\em ICLR}, 2019.

\bibitem{x-clip}
Bolin Ni, Houwen Peng, Minghao Chen, and Songyang Zhang.
\newblock Expanding language-image pretrained models for general video
  recognition.
\newblock In {\em ECCV}, 2022.

\bibitem{protuning}
Xing Nie, Bolin Ni, Jianlong Chang, Gaomeng Meng, Chunlei Huo, Zhaoxiang Zhang,
  Shiming Xiang, Qi Tian, and Chunhong Pan.
\newblock Pro-tuning: Unified prompt tuning for vision tasks.
\newblock {\em arXiv preprint: arXiv:2207.14381}, 2022.

\bibitem{st-adapter}
Junting Pan, Ziyi Lin, Xiatian Zhu, Jing Shao, and Hongsheng Li.
\newblock St-adapter: Parameter-efficient image-to-video transfer learning for
  action recognition.
\newblock In {\em NeurIPS}, 2022.

\bibitem{video-moco}
Tian Pan, Yibing Song, Tianyu Yang, Wenhao Jiang, and Wei Liu.
\newblock Videomoco: contrastive video representation learning with temporally
  adversarial examples.
\newblock In {\em CVPR}, 2021.

\bibitem{provico}
Jungin Park, Jiyoung Lee, Ig-Jae Kim, and Kwanghoon Sohn.
\newblock Probabilistic representations for video contrastive learning.
\newblock In {\em CVPR}, 2022.

\bibitem{clip}
Alec Radford, Jong~Wook Kim, Chris Hallacy, Aditya Ramesh, Gabriel Goh,
  Sandhini Agarwal, Girish Sastry, Amanda Askell, Pamela Mishkin, Jack Clark,
  Gretchen Krueger, and Ilya Sutskever.
\newblock Learning transferable visual models from natural language
  supervision.
\newblock In {\em ICLR}, 2021.

\bibitem{gpt-1}
Alec Radford, Karthik Narasimhan, Tim Salimans, and Ilya Sutskever.
\newblock Improving language understanding by generative pre-training.
\newblock Technical report, OpenAI, 2018.

\bibitem{gpt-2}
Alec Radford, Jeffrey Wu, Rewon Child, David Luan, Dario Amodei, and Ilya
  Sutskever.
\newblock Language models are unsupervised multitask learners.
\newblock Technical report, OpenAI, 2019.

\bibitem{t5}
Colin Raffel, Noam Shazeer, Adam Roberts, Katherine Lee, Sharan Narang, Michael
  Matena, Yanqi Zhou, Wei Li, and Peter~J. Liu.
\newblock Exploring the limits of transfer learning with a unified text-to-text
  transformer.
\newblock {\em J. Mach. Learn. Research}, 2020.

\bibitem{tokenlearner}
Michael~S Ryoo, AJ Piergiovanni, Anurag Arnab, Mostafa Dehghani, and Anelia
  Angelova.
\newblock Tokenlearner: Adaptive space-time tokenization for videos.
\newblock In {\em NeurIPS}, 2021.

\bibitem{laion-5b}
Christoph Schuhmann, Romain Beaumont, Richard Vencu, Cade Gordon, Ross
  Wightman, Mehdi Cherti, Theo Coombes, Aarush Katta, Clayton Mullis, Mitchell
  Wortsman, Patrick Schramowski, Srivatsa Kundurthy, Katherine Crowson, Ludwig
  Schmidt, Robert Kaczmarczyk, and Jenia Jitsev.
\newblock Laion-5b: An open large-scale dataset for training next generation
  image-text models.
\newblock In {\em NeurIPS}, 2022.

\bibitem{k400-bias}
Laura Sevilla-Lara, Shengxin Zha, Zhicheng Yan, Vedanuj Goswami, Matt Feiszli,
  and Lorenzo Torresani.
\newblock Only time can tell: Discovering temporal data for temporal modeling.
\newblock In {\em WACV}, 2021.

\bibitem{two-stream}
Karen Simonyan and Andrew Zisserman.
\newblock Two-stream convolutional networks for action recognition in videos.
\newblock In {\em NeurIPS}, 2014.

\bibitem{prompt-nlp}
Yusheng Su, Xiaozhi Wang, Yujia Qin, Chi-Min Chan, Yankai Lin, Huadong Wang,
  Kaiyue Wen, Zhiyuan Liu, Peng Li, Juanzi Li, Lei Hou, Maosong Sun, and Jie
  Zhou.
\newblock On transferability of prompt tuning for natural language processing.
\newblock In {\em NAACL}, 2022.

\bibitem{videobert}
Chen Sun, Austin Myers, Carl Vondrick, Kevin Murphy, and Cordelia Schmid.
\newblock Videobert: A joint model for video and language representation
  learning.
\newblock In {\em ICCV}, 2019.

\bibitem{pretraining1}
Chen Sun, Abhinav Shrivastava, Saurabh Singh, and Abhinav Gupta.
\newblock Revisiting unreasonable effectiveness of data in deep learning era.
\newblock In {\em ICCV}, 2017.

\bibitem{vl-adapter}
Yi-Lin Sung, Jaemin Cho, and Mohit Bansal.
\newblock Vl-adapter: Parameter-efficient transfer learning for vision-language
  tasks.
\newblock In {\em CVPR}, 2022.

\bibitem{videomae}
Zhan Tong, Yibing Song, Jue Wang, and Limin Wang.
\newblock Videomae: Masked autoencoders are data-efficient learners for
  self-supervised video pre-training.
\newblock In {\em NeurIPS}, 2022.

\bibitem{r(2+1)d}
Du Tran, Heng Wang, Lorenzo Torresani, Jamie Ray, Yann LeCun, and Manohar
  Paluri.
\newblock A closer look at spatiotemporal convolutions for action recognition.
\newblock In {\em CVPR}, 2018.

\bibitem{transformer}
Ashish Vaswani, Noam Shazeer, Niki Parmar, Jakob Uszkoreit, Llion Jones,
  Aidan~N. Gomez, Lukasz Kaiser, and Illia Polosukhin.
\newblock Attention is all you need.
\newblock In {\em NeurIPS}, 2017.

\bibitem{actionclip}
Mengmeng Wang, Jiazheng Xing, and Yong Liu.
\newblock Actionclip: A new paradigm for video action recognition.
\newblock {\em arXiv preprint: arXiv:2109.08472}, 2021.

\bibitem{beit-pretraining}
Wenhui Wang, Hangbo Bao, Li Dong, Johan Bjorck, Zhiliang Peng, Qiang Liu, Kriti
  Aggarwal, Owais~Khan Mohammed, Saksham Singhal, Subhojit Som, and Furu Wei.
\newblock Image as a foreign language: Beit pretraining for all vision and
  vision-language tasks.
\newblock {\em arXiv preprint: arXiv:2208.10442}, 2022.

\bibitem{fixed-position}
Yuqing Wang, Zhaoliang Xu, Xinlong Wang, Chunhua Shen, Baoshan Cheng, Hao Shen,
  and Huaxia Xia.
\newblock End-to-end video instance segmentation with transformers.
\newblock In {\em CVPR}, 2021.

\bibitem{p2p}
Ziyi Wang, XuminYu, YongmingRao, and Jie~Zhou JiwenLu.
\newblock P2p: Tuning pre-trained image models for point cloud analysis with
  point-to-pixel prompting.
\newblock In {\em NeurIPS}, 2022.

\bibitem{cvt}
Haiping Wu, Bin Xiao, Noel Codella, Mengchen Liu, Xiyang Dai, Lu Yuan, and Lei
  Zhang.
\newblock Cvt: Introducing convolutions to vision transformers.
\newblock In {\em ICCV}, 2021.

\bibitem{s3d}
Saining Xie, Chen Sun, Jonathan Huang, Zhuowen Tu, and Kevin Murphy.
\newblock Rethinking spatiotemporal feature learning: Speed-accuracy trade-offs
  in video classification.
\newblock In {\em ECCV}, 2018.

\bibitem{simmim}
Zhenda Xie, Zheng Zhang, Yue Cao, Yutong Lin, Jianmin Bao, Zhuliang Yao, Qi
  Dai, and Han Hu.
\newblock Simmim: A simple framework for masked image modeling.
\newblock In {\em CVPR}, 2022.

\bibitem{videoclip}
Hu Xu, Gargi Ghosh, Po-Yao Huang, Dmytro Okhonko, Armen Aghajanyan, Florian
  Metze, Luke Zettlemoyer, and Christoph Feichtenhofer.
\newblock Videoclip: Contrastive pre-training for zero-shot video-text
  understanding.
\newblock In {\em EMNLP}, 2021.

\bibitem{multiview}
Shen Yan, Xuehan Xiong, Anurag Arnab, Zhichao Lu, Mi Zhang, Chen Sun, and
  Cordelia Schmid.
\newblock Multiview transformers for video recognition.
\newblock In {\em CVPR}, 2022.

\bibitem{florence}
Lu Yuan, Dongdong Chen, Yi-Ling Chen, Noel Codella, Xiyang Dai, Jianfeng Gao,
  Houdong Hu, Xuedong Huang, Boxin Li, Chunyuan Li, Ce Liu, Mengchen Liu,
  Zicheng Liu, Yumao Lu, Yu Shi, Lijuan Wang, Jianfeng Wang, Bin Xiao, Zhen
  Xiao, Jianwei Yang, Michael Zeng, Luowei Zhou, and Pengchuan Zhang.
\newblock Florence: A new foundation model for computer vision.
\newblock {\em arXiv preprint: arXiv:2111.11432}, 2021.

\bibitem{tokens-to-tokens}
Li Yuan, Yunpeng Chen, Tao Wang, Weihao Yu, Yujun Shi, Zi-Hang Jiang,
  Francis~EH Tay, Jiashi Feng, and Shuicheng Yan.
\newblock Tokens-to-token vit: Training vision transformers from scratch on
  imagenet.
\newblock In {\em ICCV}, 2021.

\bibitem{pretraining2}
Xiaohua Zhai, Alexander Kolesnikov, Neil Houlsby, and Lucas Beyer.
\newblock Scaling vision transformers.
\newblock In {\em CVPR}, 2022.

\bibitem{vidtr}
Yanyi Zhang, Xinyu Li, Bing~Shuai Chunhui~Liu, Yi Zhu, Biagio Brattoli, Hao
  Chen, Ivan Marsic, and Joseph Tighe.
\newblock Vidtr: Video transformer without convolutions.
\newblock In {\em ICCV}, 2021.

\bibitem{randerase}
Zhun Zhong, Liang Zheng, Guoliang Kang, Shaozi Li, and Yi Yang.
\newblock Random erasing data augmentation.
\newblock In {\em AAAI}, 2020.

\end{thebibliography}
}
    \clearpage
    \newpage
    \appendix
    \setcounter{table}{0}
    \renewcommand{\thetable}{A\arabic{table}}
    \setcounter{figure}{0}
    \renewcommand{\thefigure}{A\arabic{figure}}

    \twocolumn[{%
  \renewcommand\twocolumn[1][]{#1}
    \section*{Appendix}\vskip 0.1in
  \begin{center}
    \begin{tabular}{l c c c c }
        \toprule
        Components &   K400~\cite{k400}    &  SSv2~\cite{ssv2}   & HMDB51~\cite{hmdb51}      & Diving-48~\cite{diving}  \\ 
        \midrule        
        \multicolumn{5}{l}{\hspace{-5pt}\textit{\textbf{Adapter}}} \\
        \# Adapters per block        &  4 (2 SP, 2 TP)  &  5 (2 SP, 3 TP)  &  4 (2 SP, 2 TP)  &  4 (2 SP, 2 TP)   \\
        Adapter bottleneck width        &  128  &  128  &  128  &  128  \\ \midrule
        \multicolumn{5}{l}{\hspace{-5pt}\textit{\textbf{Optimizer (AdamW~\cite{adamw}, Cosine scheduler~\cite{cosine-sched})}}} \\
        Learning rate        &  3e-4    &   5e-4    &   1e-4    &   3e-4  \\
        Weight Decay        &  5e-2    &   5e-2    &   2e-2    &   3e-2  \\
        Batch size        &  64    &   128  &   128 &   128  \\
        \midrule
        \multicolumn{5}{l}{\hspace{-5pt}\textit{\textbf{Data configuration}}} \\
        Training crop size        &  224    &   224     &   224     &      224  \\
        Frame sampling rate ($T_S$)        &  16 for $T_S=8$    &   16 for $T_S=8$  &   16 for $T_S=8$  &   16 for $T_S=8$  \\
        Frame sampling rate ($T_G$)        &  \makecell{8 for $T_G=1$ \\ 4 for $T_G=2$ \\ 2 for $T_G=3$}  & Dynamic sampling & \makecell{8 for $T_G=1$ \\ 4 for $T_G=2$ \\ 2 for $T_G=3$}  & \makecell{8 for $T_G=1$ \\ 4 for $T_G=2$ \\ 2 for $T_G=3$}\\
        RandAugment~\cite{randaugment}        &  	\checkmark  & 	\checkmark & 	\checkmark  & 	\checkmark \\
        Random erase~\cite{randerase}        &  	\xmark  & 	\checkmark & 	\checkmark  & 	\xmark \\
        \midrule
        \multicolumn{5}{l}{\hspace{-5pt}\textit{\textbf{Inference configuration}}} \\
        Testing views (temporal$\times$spatial)        &  	3 $\times$1  & 	1$\times$3 & 	2$\times$3  & 	1$\times$1 \\
        \bottomrule
        \end{tabular}
        \vspace{-5pt}
        \captionof{table}{Implementation details of \method.}\label{tab:imple_detail}
  \end{center}
}]

    In this document, we include supplementary materials for ``Dual-path Adaptation from Image to Video Transformers''.
    We first provide more concrete implementation details (\secref{sec:imple_detail}), and additional experimental results (\secref{sec:swin}), including the results using a different backbone and ablation study for the resolution of the grid-like frameset.
    Finally, we visualize more attention maps from each path to complement the effectiveness of the proposed method (\secref{sec:qual}).

\section{Implementation Details}\label{sec:imple_detail}
    We add parallel adapters in the spatial path and serial adapters in the temporal path to every transformer block.
    In our adapter, the dimension of the bottlenecked embedding is 128.
    Following prior work~\cite{adaptformer}, $\mathbf{W}_\text{down}$ is initialized with Kaiming Normal~\cite{kaiming-init} and $\mathbf{W}_{\text{up}}$ with zero initialization.
    For the SSv2~\cite{ssv2} dataset, we additionally insert one adapter before the multi-head attention layer in the temporal path for more robust temporal modeling.
    The experimental configurations according to the datasets are presented in \tabref{tab:imple_detail}.
\section{Additional Results}\label{sec:swin}

\subsection{Results with Swin-B}
    \begin{table*}[t]
    \centering
    \small
    \setlength{\tabcolsep}{7pt}
        \begin{tabular}{l c c c c c c c }
        \toprule
        Method \& Arch. &   Pretrain    &  \makecell{Model \\ \# Params}   & \makecell{Trainable \\ \# Params}      & GFLOPs    & SSv2 & HMDB51   \\ \midrule
        \rowcolor{light_gray} Full-tuning w/ Swin-B~\cite{swin}        &  IN-21K  &  88M  &  88M  &  124  & 44.3  & 61.2    \\
        ST-Adapter~\cite{st-adapter} w/ Swin-B        &  IN-21K  &  95M  &  7M  &  385 & 65.1  & -    \\ 
        \rowcolor{Light}\textbf{\method} w/ ViT-B/16        &  CLIP  &  99M  &  13M  &  642 & 69.3  & 75.6    \\
        \rowcolor{Light}\textbf{\method} w/ ViT-B/16        &  IN-21K  &  99M  &  13M  &  642  & 64.7  & 70.5    \\
        \rowcolor{Light}\textbf{\method} w/ Swin-B        &  IN-21K  &  97M  &  11M  &  287  &   67.8  &  75.2     \\
        \bottomrule
        \end{tabular}
        \vspace{-5pt}
    \caption{Performance comparisons for action recognition on the SSv2~\cite{ssv2} and HMDB51~\cite{hmdb51} dataset with different backbones and pretraining datasets.}\label{tab:add_results}
    \end{table*}
\begin{table*}[t]
    \centering
    \small
    \setlength{\tabcolsep}{7pt}{
        \begin{tabular}{lccccc}
        \toprule
        Method     & \text{\#} Frames  &  \makecell{K400 \\ R@1$\uparrow$} &   \makecell{Training \\ GPU Hours $\downarrow$}     & \makecell{Throughput \\ (V/s) $\uparrow$} & \makecell{Inference\\Latency (ms) $\downarrow$}  \\
        \midrule
        Uniformer-B~\cite{uniformer}  &  32  &   82.9  & 5000 &   3.42 &   314.58   \\
        EVL w/ ViT-B~\cite{frozen-clip}  &  8  &  82.9 & 60  & 25.53 &  102.88    \\
        \rowcolor{Light}
        \method w/ ViT-B &  16  & 85.4  & 31 &   64.21 &  15.58  \\
        \bottomrule
        \end{tabular}
        }
        \vspace{-5pt}
    \caption{Training and inference efficiency comparisons. All models are evaluated using V100-32G, following EVL~\cite{frozen-clip}.
    }\label{tab:efficiency}
    \end{table*}
\begin{table}[t]
    \centering
    \small
    \setlength{\tabcolsep}{4pt}
    \centering
        \begin{tabular}{c c c c c}
        \toprule
        \multirow{2}{*}{Resolution}  & \multicolumn{2}{c}{SSv2} & \multicolumn{2}{c}{HMDB51}    \\
        \cmidrule(lr){2-3} \cmidrule(lr){4-5}
          & {GFLOPs}   & {R@1}   & {GFLOPs}   & {R@1}    \\
        \midrule
        224$\times$224 w/ 16 frames & 642  &   69.3  & 612  &   75.6  \\
        448$\times$448 w/ 16 frames  & 846  &   70.5  & 816   & 75.8   \\
        896$\times$896 w/ 16 frames  & 1694  &   71.6  & 1632    &   76.4        \\
        224$\times$224 w/ 48 frames  & 791  &   71.2  & 778   & 76.3   \\
        \bottomrule
        \end{tabular}
        \vspace{-5pt}
        \caption{Performance comparisons for action recognition on the SSv2~\cite{ssv2} and HMDB51~\cite{hmdb51} dataset according to the resolution of the grid-like frameset.
    }\label{tab:resolution_hmdb51}
    \end{table}
    Our \method can be applied to other transformer-based pretrained image models.
    We conduct additional experiments with Swin-B~\cite{swin, swin-v2} transformer pretrained on the ImageNet-21K~\cite{imagenet}.
    The Swin-B contains 24 Swin transformer blocks with 88M parameters, requiring fewer GFLOPs than ViT-B/16~\cite{vit}.
    Each block consists of window-based and shifted window-based self-attention layers.
    As in the ViT backbones, we add parallel adapters in the spatial path and serial adapters in the temporal path to every Swin transformer block.
    Note that adapters are attached to only window-based self-attention layers while not adapting shifted window-based self-attention layers.
    For the SSv2 dataset, we use an additional adapter before the multi-head attention layer of the temporal path similar to the ViT backbones.
    The dimension of the bottlenecked embedding is set to 128.
    
    \tabref{tab:add_results} provides the experimental results of \method with Swin-B~\cite{swin, swin-v2} on the SSv2~\cite{ssv2} and HMDB51~\cite{hmdb51} datasets.
    Although the comparisons between ViT-B/16 and Swin-B backbones show the significantly low computation requirement of the Swin-B model (642 vs 287 GFLOPs with \method), we attain a comparable performance to the CLIP pretrained ViT-B/16.
    Compared to ST-Adapter~\cite{st-adapter} with Swin-B, the results consistently demonstrate the effectiveness of \method over the backbone networks, showing a higher performance of 2.7\% with Swin-B on the SSv2 benchmark.

\subsection{Additional efficiency analysis}
    We additionally compare the methods with \cite{frozen-clip, uniformer} in terms of training step time, throughput, and inference latency, following \cite{frozen-clip}.
    For a fair comparison, we obtain all results using V100-32G with PyTorch-builtin mixed precision.
    The throughput is measured with the largest batch size before out-of-memory and the inference latency is measured with a batch size of 1.
    As shown in \tabref{tab:efficiency}, \method takes about half of the training GPU hours and achieves ×2.5 more throughput and ×6.6 faster inference than EVL~\cite{frozen-clip} under the same hardware condition.

\subsection{Resolution of grid-like frameset}
    The grid-like frameset comprises a stack of 16 \textit{scaled} frames to make the same size as the original frame ($224\times 224$).
    We investigate the effectiveness of the resolution of the grid-like frameset in this section.
    Note that the impact of scaling factors that determine the temporal resolution is demonstrated in \tabref{tab:ablation} of the main paper.
    
    Specifically, we set the scaling factors $w$ and $h$ to 1, 2, and 4 while maintaining the temporal resolution as 16 such that the resolution of the grid-like frameset is $896\times 896$, $448\times 448$, and $224\times 224$, respectively.
    The backbone (ViT-B/16) is identically used and uniformly sampled 8 frames are used in the spatial path.
    Following \cite{st-adapter}, we sample one clip cropped into three different spatial views on SSv2~\cite{ssv2} (\ie, total of 3 clips) at test time.
    For HMDB51~\cite{hmdb51}, two clips sampled from a video are respectively cropped into three spatial views (\ie, a total of 6 clips).
    Since a high-resolution frameset contains more detailed information about the original frames, the highest performance is obtained with the $896\times 896$ size of the frameset in \tabref{tab:resolution_hmdb51}.
    However, the computational cost quadratically increases as the resolution of the grid-like frameset increases.
    When we use 48 frames (\ie, $T_G=3$) with the $224 \times 224$ size of the frameset, competitive performance is achieved in both datasets.
    It supports the resolution choice of \method in terms of the trade-off between performance and computational cost.

 \begin{figure*}[t]
        \centering
            \begin{subfigure}{0.45\linewidth}
            \centering
            {\includegraphics[width=0.95\linewidth]{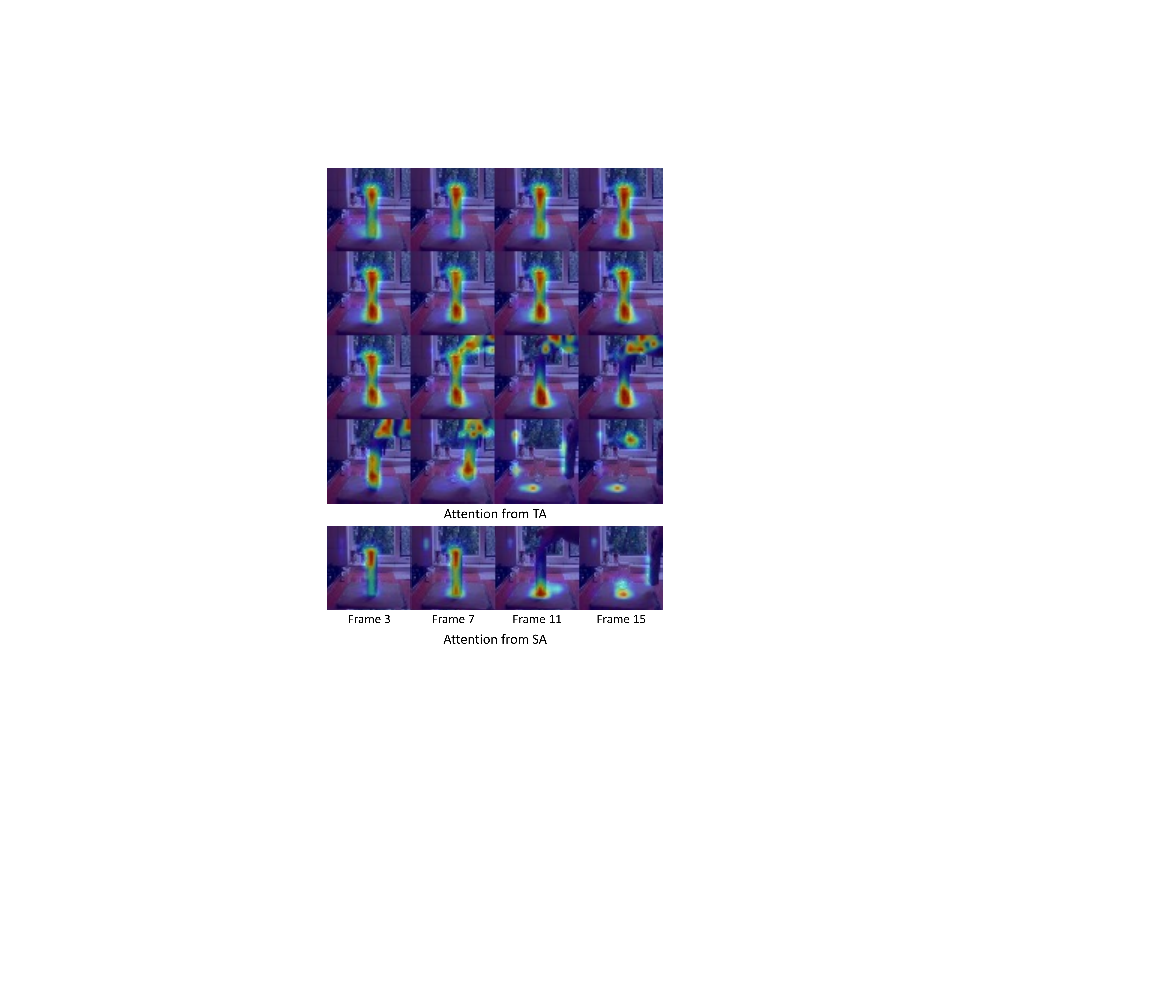}}
            \caption{Removing [something], revealing [something] behind}\label{fig:quala-supp}
           \end{subfigure}
           \begin{subfigure}{0.45\linewidth}
            \centering
            {\includegraphics[width=0.95\linewidth]{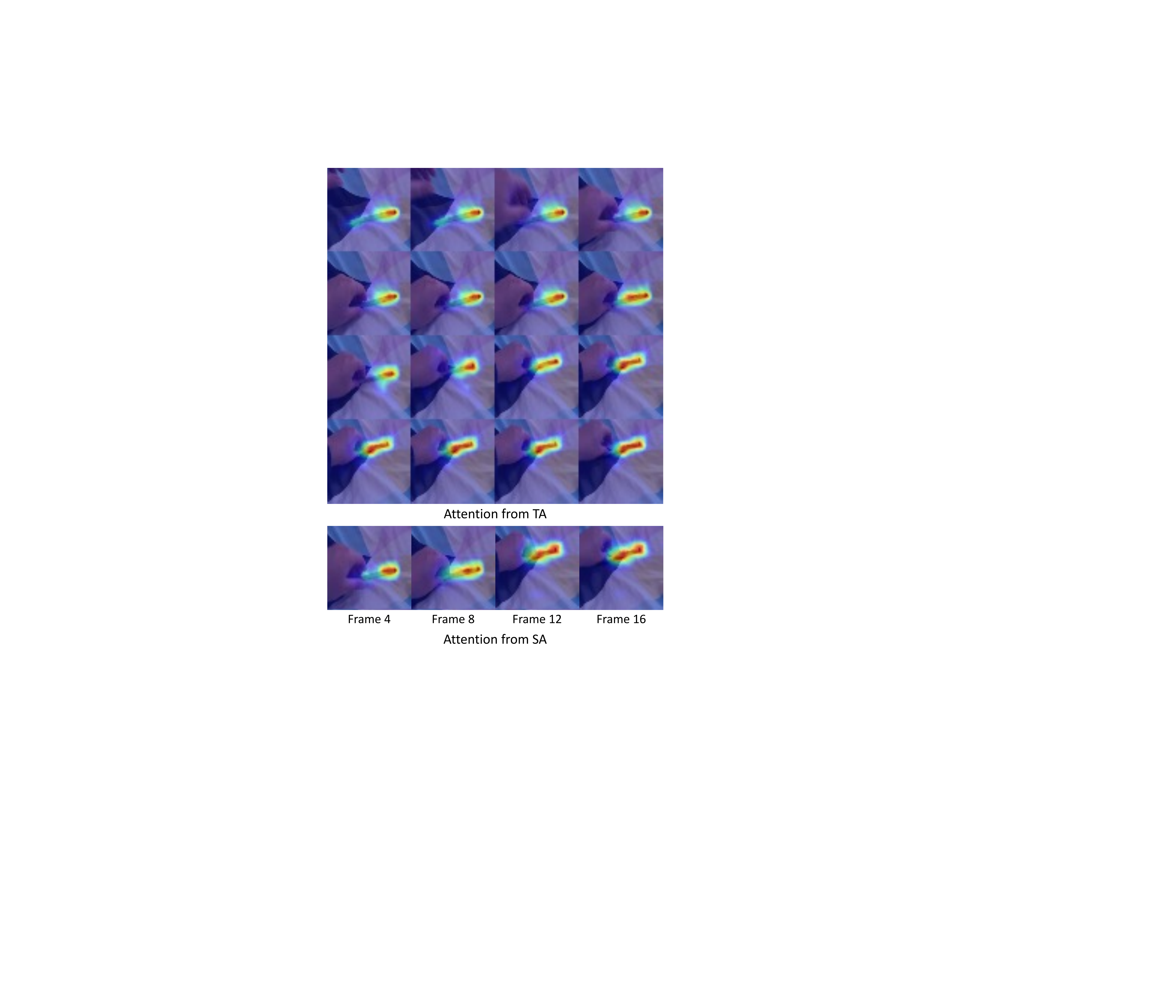}}
            \caption{Moving [something] up}\label{fig:qualb-supp}
           \end{subfigure}\\
           \begin{subfigure}{0.45\linewidth}
            \centering
            {\includegraphics[width=0.95\linewidth]{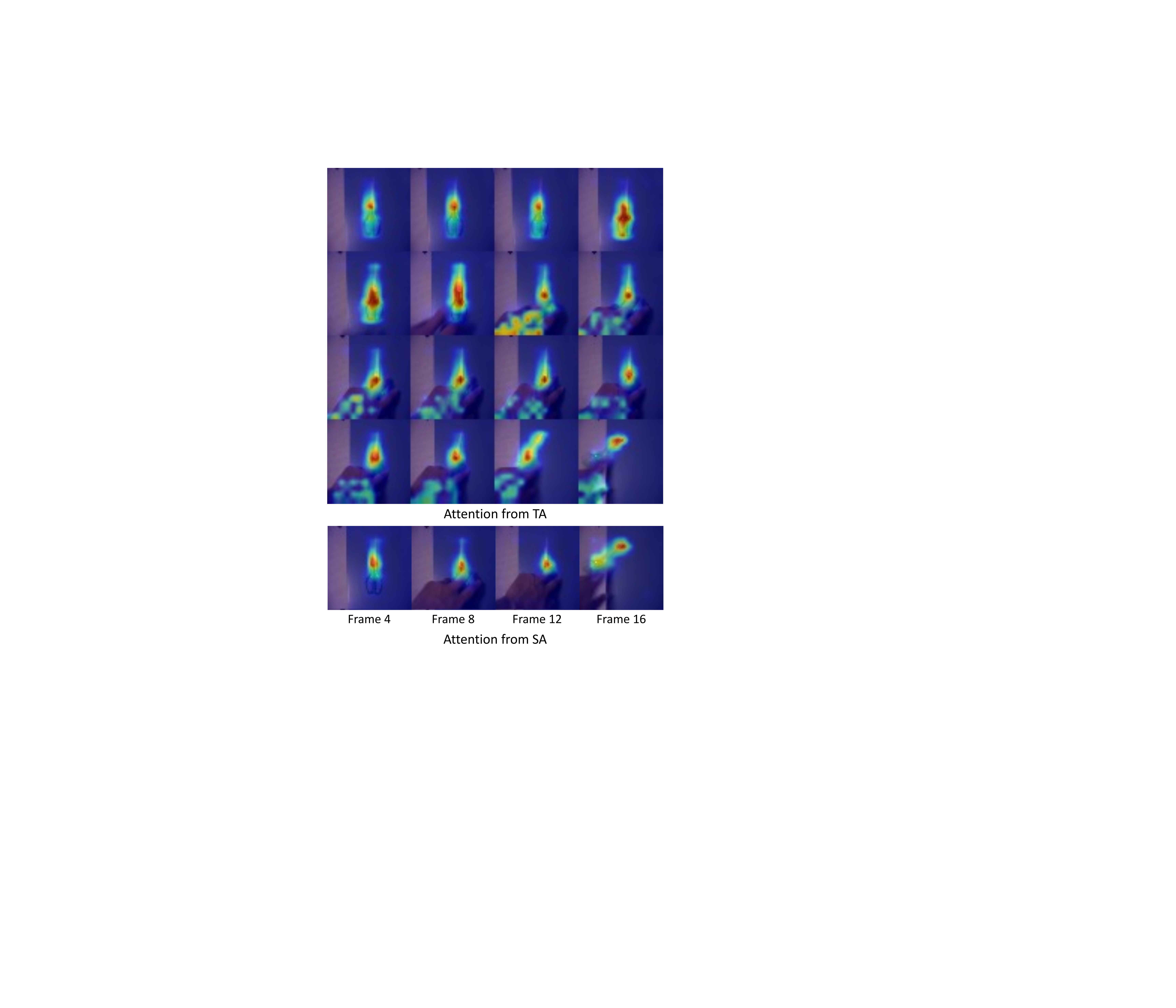}}
            \caption{Pushing [something] so that it falls off the table}\label{fig:qualc-supp}
           \end{subfigure}
           \begin{subfigure}{0.45\linewidth}
            \centering
            {\includegraphics[width=0.95\linewidth]{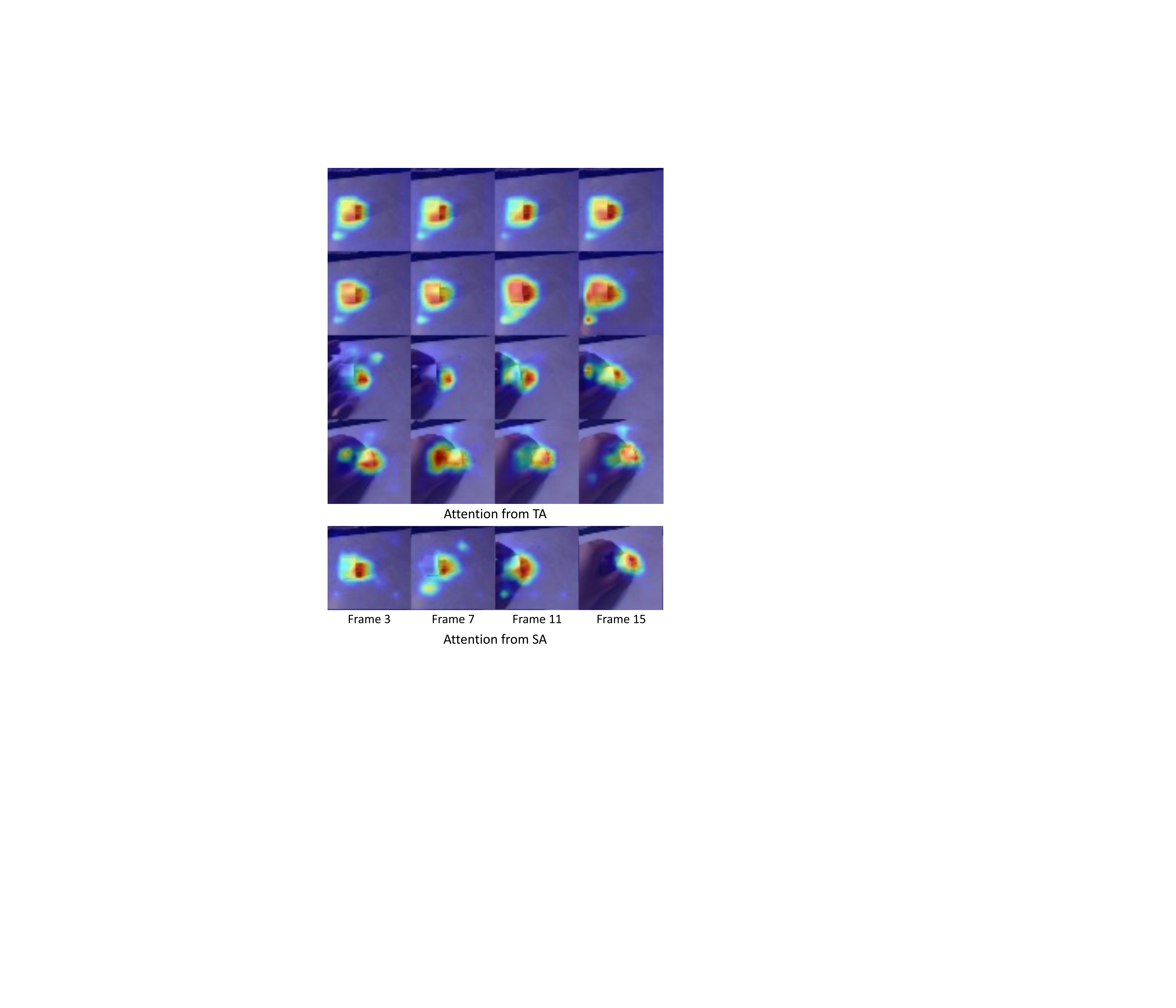}}
            \caption{Pushing [something] from left to right}\label{fig:quald-supp}
           \end{subfigure}\\
        \caption{Visualization of attention maps of each path for videos from the SSv2~\cite{ssv2} dataset.}
    \label{fig:qual_supp}
    \end{figure*}
\section{More Attention Visualization of \method}\label{sec:qual}
    The additional attention visualization is illustrated in \figref{fig:qual_supp}.
    We depict the attention maps of $\spcls$ and $\tpcls$ from the final transformer block of each path.
    All videos are sampled from the SSv2~\cite{ssv2} dataset and ViT-B/16 is used as the backbone.
    While we use 8 frames in the spatial path, the attention maps corresponding to only 4 frames are displayed for visibility.
    Interestingly, the results show that the model trained with \method is capable of focusing on dynamic action-related regions in both adaptation paths.
    As exemplified in \figref{fig:quala-supp} and \figref{fig:qualc-supp}, $\spcls$ of the spatial path tends to focus on action-related objects, and $\tpcls$ of the temporal path concentrates on action-related movements.
    Therefore, the two paths complement each other, leading to spatiotemporal modeling.
    \newpage
%-------------------------------------------------------------------------

\clearpage

\end{document}